\documentclass{article}

\makeatletter
\def\input@path{{neurips2026/}{sty/}}
\makeatother
\PassOptionsToPackage{numbers,sort&compress}{natbib}
\usepackage[final,main]{neurips_2026}

\usepackage[utf8]{inputenc}
\usepackage[T1]{fontenc}
\usepackage{hyperref}
\usepackage{url}
\usepackage{booktabs}
\usepackage{longtable}
\usepackage{amsfonts}
\usepackage{amsmath}
\usepackage{amssymb}
\usepackage{nicefrac}
\usepackage{microtype}
\usepackage{xcolor}
\usepackage[table]{colortbl}
\definecolor{multigold}{RGB}{184,134,11}
\newcommand{\win}[1]{\textcolor{multigold}{\bfseries\boldmath #1}}
\usepackage{graphicx}
\usepackage{caption}
\usepackage{tikz}
\usetikzlibrary{arrows.meta,positioning,calc,fit,backgrounds,decorations.pathreplacing}

\title{Every Measurement, Every Direction, All at Once:\\Multimodal Flow Matching for Molecules and Spectra}

\author{%
  Thorben Prein\thanks{Equal contribution.} \\
  Lila Sciences \\
  \texttt{tprein@lila.ai} \\
  \And
  Elton Pan\footnotemark[1] \\
  Lila Sciences \\
  \texttt{epan@lila.ai} \\
  \And
  Rafael Gómez-Bombarelli \\
  Lila Sciences \\
  \texttt{rgbombarelli@lila.ai} \\
  \And
  Santiago Miret \\
  Lila Sciences \\
  \texttt{smiret@lila.ai} \\
}

\begin{document}

\maketitle

\begin{abstract}
Generative modeling in science is often data-limited, where experiments are expensive. However, each experiment may yield several coupled measurements of the same physical entity. Standard  generative models often train only on a single direction, discarding supervision from the rest. We argue that each experiment contains more training signal than this asymmetry allows the model to exploit: random masking enables measurements to serve as either conditioning inputs or prediction targets, yielding multiple supervised denoising tasks from a single experiment. We investigate this premise with MOSAIC, a multimodal flow matching transformer over molecular 3D structure and UV, IR, and Raman spectra. MOSAIC is a \textit{single} model that is capable of \textit{any-to-any} generation: generating any subset of modalities from any disjoint subset of the others. On QM9S, MOSAIC achieves 88.8\% molecular graph recovery, outperforming the strongest spectra-to-structure baseline by 23.2\% in structure generation, while simultaneously outperforming a structure-to-spectra specialist model in spectral generation. Ablations show that the any-to-any objective improves structure generation over structure-only training across model sizes. Furthermore, MOSAIC supports arbitrary subsets of available measurements and enables molecular identification by matching generated spectra to experimental references.
\end{abstract}

\begin{figure}[h]
\centering
\includegraphics[width=\linewidth]{"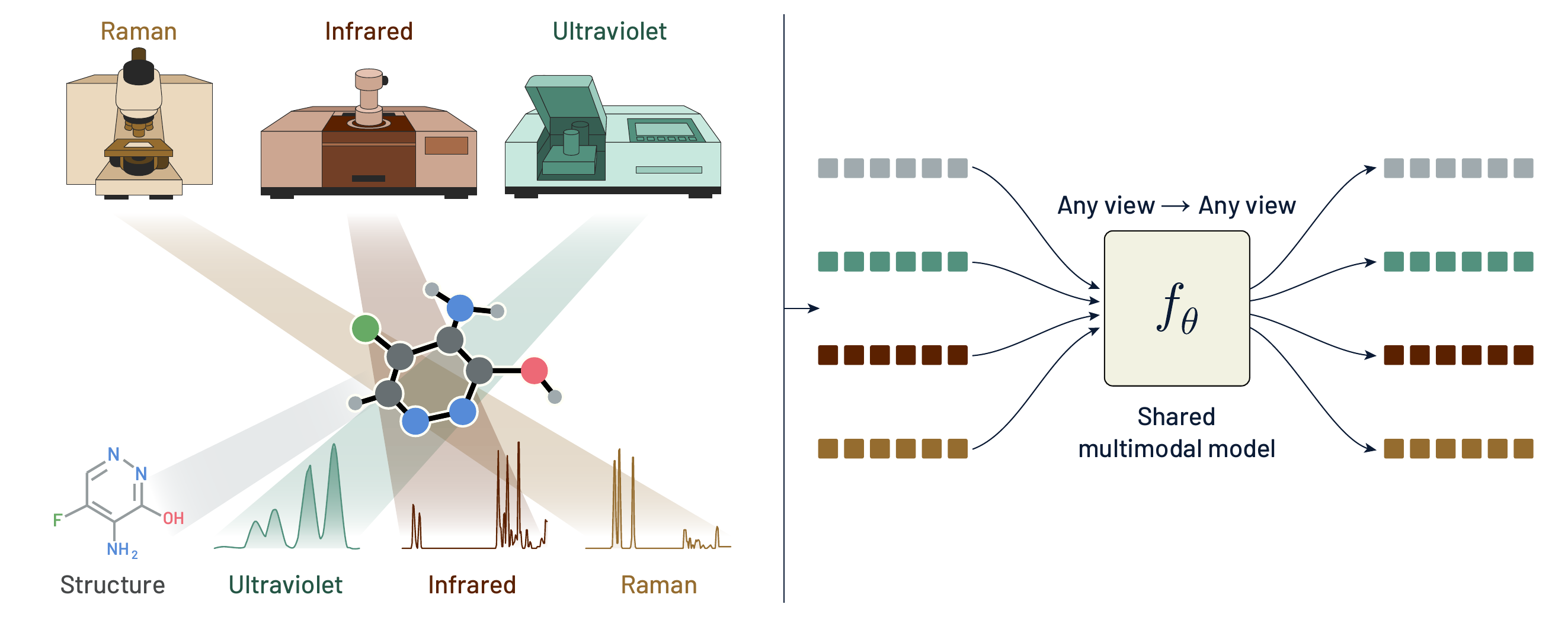"}
\caption{Scientific measurements provide complementary views of the same physical entity. With multimodal \textit{any-to-any} training, a single model learns to generate any subset of modalities from any disjoint subset.}
\label{fig:measurement_views}
\end{figure}

\begin{figure}[t]
\centering
\includegraphics[width=\linewidth]{"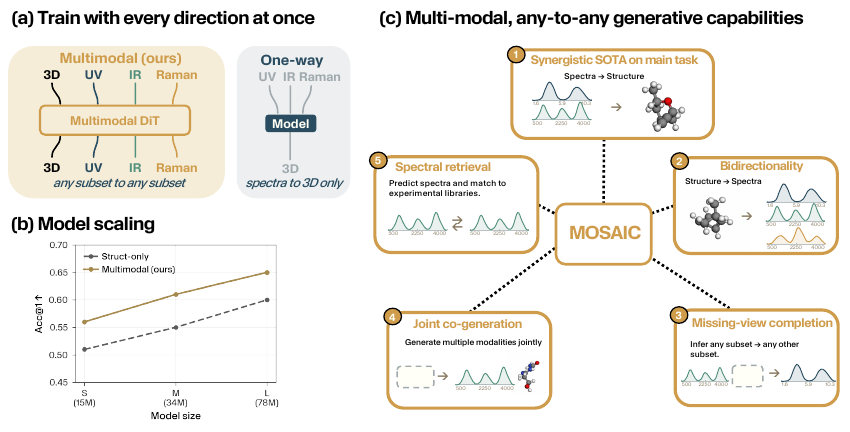"}
\caption{\textbf{MOSAIC: Multimodal any-to-any training unlocks new capabilities in scientific generative models.}
  \textbf{(a)}~Joint training over molecular structure, UV, IR, and Raman spectra with sampled conditioning and target modalities.
  \textbf{(b)}~Structure generation accuracy improves with model size and exceeds that of the structure-only variant.
  \textbf{(c)}~Any-to-any training enables structure generation, spectrum generation, missing-view completion, joint generation and spectral retrieval with a single model.}
\label{fig:hero}
\end{figure}

\section{Introduction}
\label{sec:intro}

In scientific domains, experiments can be costly but often provide several complementary sources of information: a molecule pairs a three-dimensional structure with simulated UV, IR, and Raman spectra (Figure \ref{fig:measurement_views}, left)~\citep{qm9_2014,zou2023qm9s}, each offering a distinct view of the same physical entity~\citep{wang2023aiforscience}. These views are physically linked, molecular structure governs the spectral response, while UV, IR, and Raman spectra constrain the structures consistent with them. Standard conditional generative models typically learn only one direction of this relationship, such as generating molecular structures from spectra.
We argue that this asymmetry arises from task setup rather than the data itself: When scientific measurements are coupled by a shared underlying system and co-occur at training time, every view carries supervision about that system ~\citep{huh2024platonic,balestriero2026platonic}. \textit{Generative models that impose a constraint on the direction(s) of generation potentially leave untapped supervision signal on the table.}

Consequently, we propose that multimodal generative models should be trained with no such constraint as shown in Figure~\ref{fig:hero}a. Every modality can act as either a target or condition. Therefore, one expensive sample yields multiple supervised denoising tasks for the same model. Our approach significantly expands the model's inference-time capabilities (Figure~\ref{fig:hero}c). It can perform inference across arbitrary observed and target modality subsets, unlocking capabilities such as matching to spectral libraries, missing-modality imputation, structure-to-spectra generation, and cross-spectra translation.

Building on this principle, we introduce MOSAIC, a multimodal flow matching model for any-to-any generation across molecular structures and spectra. MOSAIC encodes molecular structures into per-atom latent representations using a UAE-3D autoencoder~\citep{uae3d2025}, while representing UV, IR, and Raman spectra as sequences of spectral tokens. These representations are jointly modeled by a single non-equivariant MMDiT~\citep{sd32024}, trained with a rectified-flow objective to denoise any subset of the four modalities. Unlike the equivariant denoiser used in prior work~\citep{diffspectra2025}, our model operates on learned structural representations. On the QM9S dataset, we observe a 5\% multimodal advantage over unimodal generation in structural elucidation (Figure~\ref{fig:hero}b). The best MOSAIC configuration achieves $88.8$\% structural recovery on QM9S, outperforming the best baseline by over $20$\%, while achieving faster inference. Beyond spectra-to-structure generation, MOSAIC also supports structural elucidation through matching in spectral space. We demonstrate this reverse direction through structure-conditioned retrieval against experimental IR spectra (Section~\ref{sec:exp:struct2ir}). Together, these results illustrate how any-to-any training can improve structure generation while expanding the range of tasks supported by a single model. Our contributions are:

\begin{enumerate}
    \item We present MOSAIC, a flow matching model trained on any-to-any generation (Figure~\ref{fig:method}a) for multimodal scientific data, where random target/conditioning masks enable every modality to serve as both target and conditioning under a single unified model. We demonstrate that MOSAIC achieves state-of-the-art performance on the spectra-to-structure task, with faster inference.
    \item We show that MOSAIC unlocks new capabilities, like bidirectional generation and missing-modality completion within a single model (Figure~\ref{fig:crossmodal}). We demonstrate how this enables structure-conditioned retrieval against experimental IR spectra (Section~\ref{sec:exp:struct2ir}).
    \item We study how MOSAIC scales with model size and training data (Figure~\ref{fig:scaling}). Linear probing further suggests that learning jointly from structures and spectra yields richer molecular representations than structure-only training (Figure~\ref{fig:probe}).
\end{enumerate}

\definecolor{capyes}{RGB}{215,235,217}
\definecolor{regimems}{RGB}{222,233,247}
\definecolor{regimeir}{RGB}{249,236,196}
\begin{table}[b]
  \centering
  \caption{Capabilities of selected spectra-conditioned molecular generators.
    Any-to-any denotes one model supporting arbitrary disjoint input/output modality subsets.
    Capabilities refer to the primary generator.}
  \label{tab:related}
  \small
  \setlength{\tabcolsep}{5pt}
  \renewcommand{\arraystretch}{1.2}
  \begin{tabular}{@{}lllccc@{}}
    \toprule
    Method
      & Spectral inputs
      & Generation
      & \shortstack{Gen.\\3D}
      & \shortstack{Gen.\\spectra}
      & \shortstack{Any-to-\\any} \\
    \midrule
    DiffMS \citep{diffms2025} & \cellcolor{regimems}MS/MS & discrete diffusion & --- & --- & --- \\
    FlowMS \citep{flowms2026} & \cellcolor{regimems}MS/MS & discrete flow matching & --- & --- & --- \\
    FRIGID \citep{frigid2026} & \cellcolor{regimems}MS/MS & masked diffusion & --- & --- & --- \\
    Alberts et al.~\citep{alberts2024irstructure} & \cellcolor{regimeir}IR & autoregressive & --- & --- & --- \\
    SMEN--SD~\citep{spectra2structure2024} & \cellcolor{regimeir}IR & autoregressive & --- & --- & --- \\
    SpectraLLM~\citep{spectrallm2026} & \cellcolor{regimeir}UV, IR, Raman, NMR, MS & autoregressive & --- & --- & --- \\
    DiffSpectra \citep{diffspectra2025} & \cellcolor{regimeir}UV, IR, Raman & continuous diffusion & \cellcolor{capyes}\checkmark & --- & --- \\
    \midrule
    \textbf{MOSAIC (Ours)} & \cellcolor{regimeir}UV, IR, Raman & flow matching & \cellcolor{capyes}\checkmark & \cellcolor{capyes}\checkmark & \cellcolor{capyes}\checkmark \\
    \bottomrule
  \end{tabular}
\end{table}

\section{Related work}
\label{sec:related}

\paragraph{Generative models for molecules.}
Diffusion~\citep{song2019score,ho2020ddpm,karras2022edm} and flow
matching~\citep{flowmatching,rectifiedflow2022}  underpin many recent approaches on generative models for  molecular graphs and 3D structures.
A central challenge is to model categorical chemical features alongside
continuous geometry while respecting their different symmetries.
Graph generation can be formulated as discrete denoising over atom and
bond types~\citep{digress2023}, whereas generation in 3D can jointly
denoise atom types and coordinates with spatially equivariant
networks~\citep{edm2022}.
Joint graph-and-coordinate models combine discrete and continuous
diffusion to learn connectivity and geometry together~\citep{midi2023}.
Flow matching offers a complementary formulation, learning continuous
transport toward molecular structures ~\citep{song2023equifm,dunn2024flowmol}.
An alternative approach is to
learn a latent representation, separating molecular reconstruction from
generative modeling.
The choice of latent space then determines how chemistry and geometry
enter the generator.
GeoLDM~\citep{geoldm2023} retains geometric structure through invariant
scalar and equivariant geometric features, using equivariant networks
for encoding, decoding, and latent denoising. Bonds are inferred from
decoded atom types and interatomic distances.
UAE-3D~\citep{uae3d2025} instead encodes atom types, bond types, and
coordinates into a unified sequence of continuous latent tokens.
Rotation and translation augmentation encourage SE(3)-equivariant
reconstruction without imposing spatial equivariance architecturally,
and its UDM-3D generator models the latents with a standard DiT.
This representation provides a natural interface for combining molecular
structure with other continuous modalities.
We adopt it in an MMDiT-based generator~\citep{sd32024} trained with flow
matching to jointly model molecular structure and spectra.

\paragraph{Multimodal generation in language and vision.}
Three complementary lines of work are relevant to our approach.
The first focuses on text-conditioned generation, including text-to-image
\citep{nichol2022glide,ramesh2022dalle2,rombach2022ldm,saharia2022imagen,sd32024},
text-to-audio
\citep{kreuk2023audiogen,liu2023audioldm,copet2023musicgen,evans2024stableaudio},
and text-to-video
\citep{ho2022imagenvideo,singer2023makeavideo,blattmann2023videoldm,blattmann2023svd}
using autoregressive, diffusion, or flow matching models.
Related text-to-3D and 4D methods
\citep{poole2023dreamfusion,lin2023magic3d,bahmani20234dfy}
optimize scene representations using pretrained diffusion models as priors.
Across these settings, text serves as the conditioning modality for
non-text outputs.
A second line supports multiple generation directions and, in some cases,
interleaved multimodal outputs.
Shared-transformer approaches span distinct training objectives:
CM3Leon~\citep{cm3leon2023} tokenizes images and text into a shared sequence
and trains a decoder-only transformer with a causal-infilling objective,
Chameleon~\citep{chameleon2024} scales early-fusion autoregressive modeling
of interleaved image and text tokens.
Unified-IO~2~\citep{lu2024unifiedio2} unifies vision, language, audio, and
action with a shared autoregressive encoder--decoder transformer.
Transfusion~\citep{zhou2024transfusion} combines next-token prediction for
text with diffusion over continuous image latent patches in one transformer,
while Show-o~\citep{xie2025showo} combines autoregressive text prediction
with discrete image diffusion.
Diffusion models also support flexible multimodal generation,
UniDiffuser~\citep{bao2023unidiffuser} models marginal, conditional, and
joint image--text distributions with one backbone using per-modality noise
levels, while CoDi~\citep{tang2023codi} composes modality-specific latent
diffusion models through shared-space conditioning and cross-attention
between diffusers.
A third line uses masked multimodal pre-training:
MultiMAE~\citep{bachmann2022multimae} reconstructs masked patches across
modalities, while 4M~\citep{mizrahi20234m} and
4M-21~\citep{bachmann20244m21} predict discrete target tokens from sampled
visible tokens across modalities.
We draw on their principle of predicting missing multimodal information from observed inputs. MOSAIC samples whole scientific modalities as clean conditions or noisy targets and learns continuous flow matching velocities, supporting conditional and joint generation across molecular structure and spectra.

\paragraph{Structural elucidation from spectra.}

Molecular structure elucidation from spectra is an inverse problem:
different structures can produce similar spectral observations.
Generative models address this ambiguity by modeling distributions
over candidate structures conditioned on spectra. Existing approaches
include autoregressive, diffusion, and flow matching models. Autoregressive approaches treat the structure as a SMILES string
and learn next-token prediction conditioned on spectra:
Alberts et al.~\citep{alberts2024irstructure} train a transformer to map
IR spectra and molecular formulae to SMILES strings, Kanakala et
al.~\citep{spectra2structure2024} train a CLIP-style contrastive model
with an EGNN molecule encoder and a ViT-style spectra encoder for
IR-to-molecule retrieval, paired with a GPT-style SMILES decoder, and
SpectraLLM~\citep{spectrallm2026} represents spectra as peak-extracted
natural-language descriptions and finetunes a Qwen3-32B with LoRA
for end-to-end structure prediction across IR, Raman, UV-Vis, NMR, and
MS. Diffusion and flow matching approaches instead model the
structure through iterative refinement, applied either to molecular
graphs and 3D coordinates or to discrete molecular-string tokens such as SAFE~\citep{noutahi2024gotta}.
The closest 3D analogue is DiffSpectra~\citep{diffspectra2025}, which performs
conditional graph-and-coordinate diffusion using an SE(3)-equivariant DMT and a
spectral encoder pretrained with masked reconstruction and contrastive
alignment with a pretrained 3D molecular encoder. In the
mass-spectra setting, DiffMS~\citep{diffms2025} couples a learned spectral
encoder to a formula-conditioned graph diffusion decoder,
FlowMS~\citep{flowms2026} replaces that diffusion decoder with flow
matching, and FRIGID~\citep{frigid2026} scales a diffusion-language-model
spectra-to-structure approach over SAFE molecular strings. Prior approaches for generating molecular structures
from spectra are shown in Table~\ref{tab:related}. 
MOSAIC extends this setting to \textit{any-to-any} generation within a single model, supporting spectra-to-structure, structure-to-spectra, and generation across arbitrary disjoint subsets of conditioning and target modalities (Figure~\ref{fig:method}a).

\section{Method}
\label{sec:method}

\begin{figure}[t]
\centering
\includegraphics[width=\linewidth]{"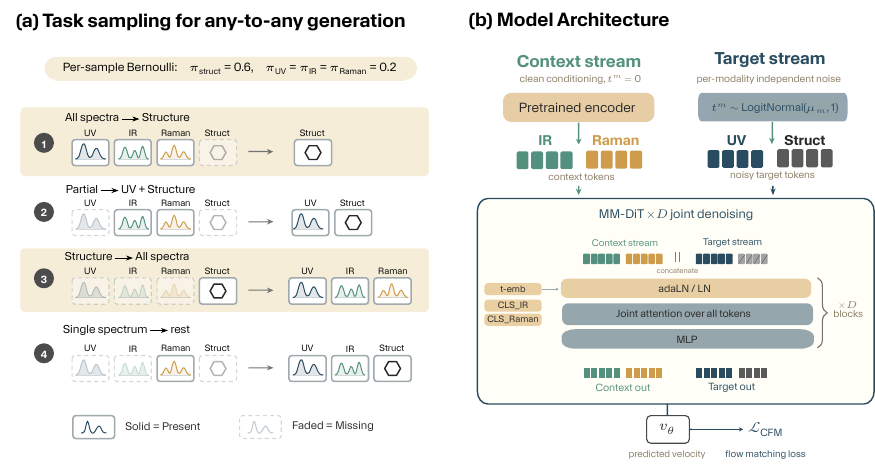"}
\caption{\textbf{Overview of MOSAIC training and architecture.}
\textbf{(a)} At each training step, we perform \textit{any-to-any} training by randomly selecting target modalities $\mathcal{T}$ and conditioning (context) modalities $\mathcal{C}$. For example, IR and Raman provide conditioning for generating UV and molecular structure. \textbf{(b)} The pretrained encoder converts the conditioning modalities into context tokens, while each target modality is noised to $x_t^m$ at timestep $t^m$. We pass both sets of tokens to an MMDiT \citep{sd32024}, where each of the $D$ blocks applies joint self-attention across them. The model predicts target velocities $v_\theta$, which are used at inference to iteratively generate the target modalities from noise. The same model handles different combinations of conditioning and target modalities.}
\label{fig:method}
\end{figure}

MOSAIC is a multimodal flow matching transformer over four coupled views of one molecule: structure, UV, IR, and Raman (Figure~\ref{fig:method}a). Each training sample is a tuple $\mathbf{x} = (x^\textrm{struct}, x^\textrm{UV}, x^\textrm{IR}, x^\textrm{Raman})$ over the modality set $\mathcal{M} = \{\textrm{struct}, \textrm{UV}, \textrm{IR}, \textrm{Raman}\}$, and any conditional generation task selects a target subset $\mathcal{T}$ and a conditioning subset $\mathcal{C} \subseteq \mathcal{M} \setminus \mathcal{T}$. Molecular structures are tokenized per atom via a UAE-3D \citep{uae3d2025} autoencoder, decoding to atom and bond types, as well as 3D coordinates. The spectral modalities UV, IR, and Raman are split into 1-D patches.

We train MOSAIC in two stages: encoder pretraining and generative training. During encoder pretraining, we train a shared encoder with self-supervised objectives across all four modalities, including CLIP-style contrastive alignment of representations from the same molecule (Section~\ref{sec:method:phase1}). During generative training, we train a non-equivariant MMDiT generator \citep{sd32024} that applies joint self-attention over structure and spectral tokens (Section~\ref{sec:method:phase2}). This architecture brings conditioning and target modalities together in a single multimodal token sequence. During training, we sample target modalities using per-modality Bernoulli distributions, with the remaining modalities providing conditioning. The resulting masks $(\mathcal{T}, \mathcal{C})$ allow spectra-to-structure, structure-to-spectra, and joint co-generation to share one set of weights and one objective (Figure~\ref{fig:method}a).

\subsection{Multimodal encoder pretraining}
\label{sec:method:phase1}

During encoder pretraining, we train a shared transformer to align representations across scientific modalities and provide conditioning features for generative training. We initialize the transformer with eight blocks and width $512$ over spectral patches, UAE-3D atom latents, and one CLS token per modality. Spectra are intensity arrays of different lengths, with patches projected to the shared width. Joint self-attention contextualizes modalities, while each spectrum retains its own physical grid (Appendix~\ref{sec:app:phase1}). We train this encoder with three complementary self-supervised objectives under modality dropout and per-token masking.

\emph{Masked reconstruction} recovers normalized spectral patches
and UAE-3D structure latents at masked positions using
modality-specific prediction heads. We minimize mean squared
error, with a larger weight on structure reconstruction. \emph{Cross-modal contrastive alignment} uses symmetric InfoNCE
\citep{infonce2018}, averaged over all $\binom{4}{2}\!=\!6$ pairs
of modality-specific CLS embeddings. For each modality pair,
embeddings from the same molecule form a positive pair, while
embeddings from different molecules in the batch serve as negatives. \emph{Self-distillation} supervises the student's early-layer
representations using later-layer representations from an
exponential moving average (EMA) teacher. We mean-pool each
full token sequence, project and L2-normalize the pooled
representations, and minimize a scaled cosine-distance loss
between the student and teacher representations. The overall objective is a weighted sum of these three losses
(Appendix~\ref{sec:app:phase1}).

\subsection{Multimodal any-to-any flow matching}
\label{sec:method:phase2}

For the flow matching model (decoder), we train a non-equivariant MMDiT \citep{sd32024,dit2023} with rectified flow \citep{flowmatching,rectifiedflow2022,sd32024}, operating on standardized UAE-3D structure latents and normalized spectral patches. Each MMDiT block applies joint self-attention to noisy target tokens and encoder context tokens, using separate QKV projections, adaLN modulation, and MLPs for the two streams (Figure~\ref{fig:method}b). Before encoding, target modality inputs are replaced by learned mask embeddings while retaining their token positions. The pretrained encoder is finetuned jointly with the MMDiT and supplies its full token sequence to the attention operation. Its four CLS states are also concatenated and linearly projected for global adaLN modulation \citep{dit2023}.

Conditions and targets are selected as whole modalities, without token subsampling (Appendix~\ref{sec:app:phase1}). For each training example, we sample target modalities, corrupt their representations, and regress the corresponding velocities. \emph{(1)~Target-set sampling}: the target subset $\mathcal{T}$ is drawn from independent per-modality Bernoullis biased toward structure as the main downstream target, with the remaining modalities providing context before classifier-free dropout (Figure~\ref{fig:method}a).

\begin{equation}
  z_m \sim \mathrm{Bernoulli}(\pi_m), \quad
  \widetilde{\mathcal{T}}=\{m\in\mathcal{M}:z_m=1\}.
\label{eq:mask}
\end{equation}

where $\pi_m$ are per-modality target-sampling probabilities. We set $\mathcal{T}=\widetilde{\mathcal{T}}$ when nonempty, and $\mathcal{T}=\{\textrm{struct}\}$ otherwise, then define $\mathcal{C}=\mathcal{M}\setminus\mathcal{T}$. \emph{(2)~Per-modality noising}: targets are noised at independent timesteps $t^m \sim \mathrm{LogitNormal}(\mu_m, 1)$ while conditioning modalities pass clean through the pretrained encoder,
\begin{equation}
\begin{aligned}
  x_t^m &= (1 - t^m)\,x_0^m + t^m\,\varepsilon^m,
   \quad \varepsilon^m \sim \mathcal{N}(0, I)
   && \text{for } m \in \mathcal{T}, \\
  x_t^m &= x_0^m
   && \text{for } m \in \mathcal{C}.
\end{aligned}
\label{eq:permod-noise}
\end{equation}
Here $x_0^m$ denotes clean normalized spectral patches or standardized UAE-3D structure latents. The modality-specific logit-space means $\mu_m$ control the training noise distributions, with $\mu_\textrm{struct}=0$ and $\mu_\textrm{UV}=\mu_\textrm{IR}=\mu_\textrm{Raman}=2$, so spectral targets are sampled at higher noise levels more frequently than structure targets (Appendix~\ref{sec:app:timestep_mu}). \emph{(3)~Velocity regression}: the model regresses the rectified-flow velocity $\varepsilon^m-x_0^m$ per target,

\begin{equation}
  \mathcal{L}_{\mathrm{gen}} = \sum_{m\in\mathcal{M}} w_m\,\ell_m
    + \lambda_\mathrm{ce}\,\mathcal{L}_\mathrm{atom\text{-}CE}.
\label{eq:p2}
\end{equation}

Here $\ell_m$ is the velocity MSE averaged over features and valid target tokens, $w_m$ weights each modality, and $\mathcal{L}_\mathrm{atom\text{-}CE}$ averages atom-type cross-entropy over valid structure-target atoms. To compute this term, we undo standardization on the predicted clean structure latents, $\hat{x}_0=x_t-tv_\theta$, and pass them through the frozen UAE-3D decoder. Appendix~\ref{sec:app:phase2} describes this in detail.

Sampling different target sets trains a shared generator for structure generation, spectrum generation, and joint generation. We use classifier-free guidance (CFG) \citep{cfg2022}, jointly masking the encoder context tokens and CLS states for dropped examples during training. All generative training hyperparameters and the inference procedure are reported in Appendix~\ref{sec:app:phase2}. \label{sec:method:variable_count}

\begin{figure}[!t]
  \centering
  \includegraphics[width=\linewidth]{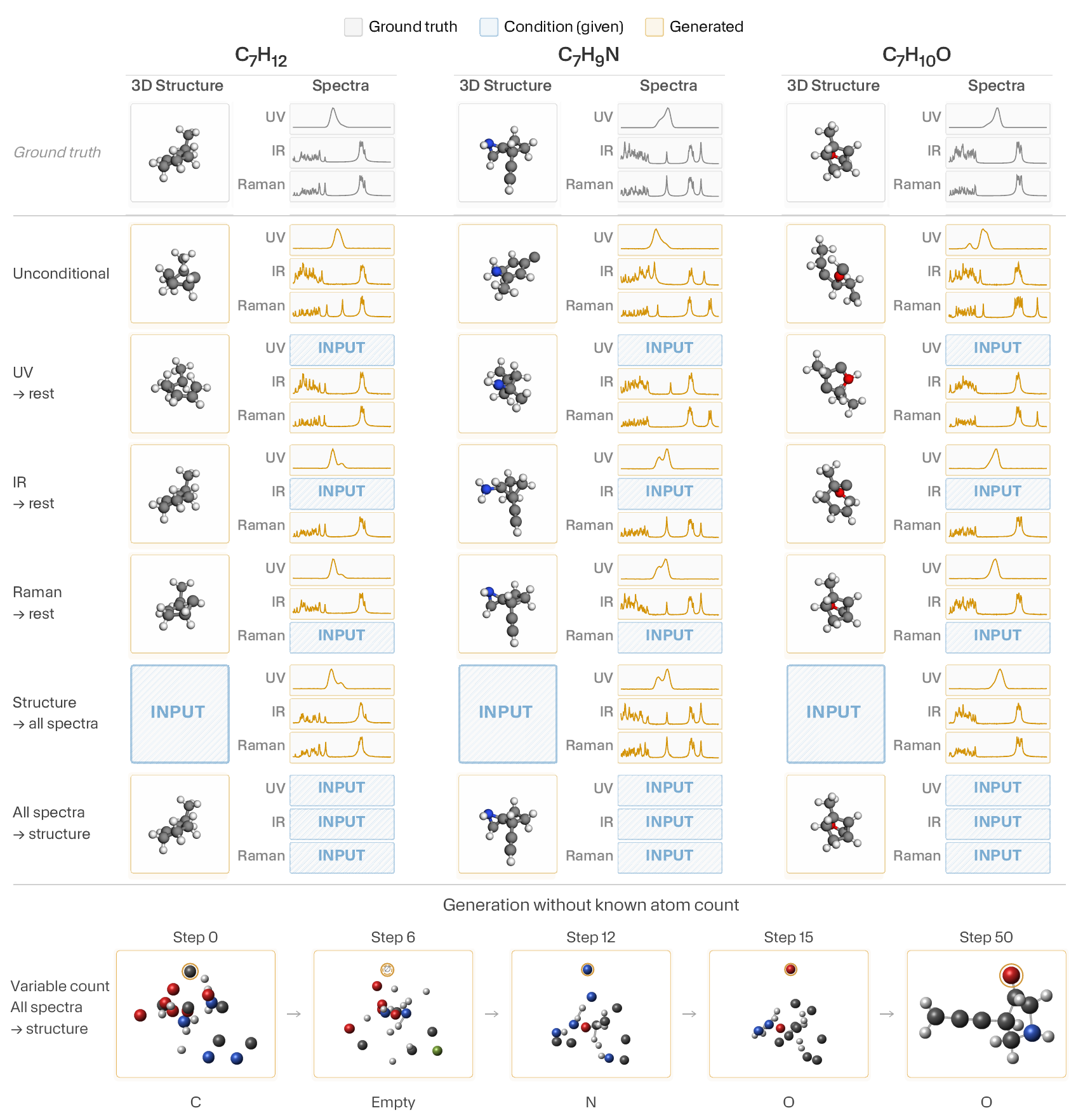}
  \caption{\textbf{Multimodal any-to-any generation with MOSAIC.} \textbf{Top:} a single MOSAIC model handles six conditioning settings for three test molecules, including spectra-to-structure, structure-to-spectra, partial completion, and unconditional joint generation. Blue denotes given modalities and gold denotes generated outputs. \textbf{Bottom:} a separate variable-count variant infers atom occupancy from spectra using a fixed latent slot budget. The highlighted atom changes from carbon (black) to empty, then nitrogen (blue) and oxygen (red) during denoising.}
  \label{fig:crossmodal}
\end{figure}

Our main spectra-to-structure protocol uses the ground-truth atom count to initialize the active structure tokens. To accommodate scenarios where ground-truth atom count is not known a priori, we train a variant that generates molecular structures with variable atom count by denoising structure latents over a fixed maximum number of slots.
This variant uses a randomly initialized conditioning encoder and additional atom- and bond-supervision losses, as detailed in Appendix~\ref{sec:app:phase2}.
The atom decoder assigns each slot either an atom type or
an empty class, allowing the model to infer the atom count. Figure~\ref{fig:crossmodal} (bottom)
illustrates how slot occupancy and atom identity evolve
during denoising. Appendices~\ref{sec:app:phase2}
and~\ref{sec:app:eval} provide training and decoding details.

\section{Experiments}
\label{sec:exp}

\newcommand{\best}[1]{\win{#1}}
\begin{table}[t]
  \centering
    \caption{\textbf{Multimodal generation on QM9S}; ``All spec.''\ uses UV+IR+Raman.
  Reported times are per-molecule averages from batched inference.
  $^{\dagger}$With molecular formula and beam search ($n{=}10$).
  $^{\ddagger}$Inference-time single-modality ablations of the
same all-spectra DiffSpectra checkpoint, with the other two
spectral inputs set to zero and no additional training.}
  \label{tab:headline}
  \small
  \setlength{\tabcolsep}{2pt}
  \resizebox{\linewidth}{!}{%
  \begin{tabular}{@{}llcccccccccc@{}}
    \toprule
    & Cond.
      & Acc@1{\scriptsize$\,\uparrow$}
      & H-RMSD{\scriptsize$\,\downarrow$}
      & Comp{\scriptsize$\,\uparrow$}
      & Tani\textsubscript{MG}{\scriptsize$\,\uparrow$}
      & Valid{\scriptsize$\,\uparrow$}
      & \multicolumn{3}{c}{Spec cos{\scriptsize$\,\uparrow$}}
      & Steps
      & ms/mol{\scriptsize$\,\downarrow$} \\
    \cmidrule(lr){8-10}
    & & & & & & & UV & IR & Raman & & \\
    \midrule
    \multicolumn{12}{@{}l}{\textit{Autoregressive spectra-to-SMILES}} \\[2pt]
    \quad Alberts et al.~\citep{alberts2024irstructure} & IR+form. & $53.7^{\dagger}$ & --- & $94.3^{\dagger}$ & $0.726^{\dagger}$ & $98.5^{\dagger}$ & --- & --- & --- & --- & $19$ \\
    \quad SMEN--SD~\citep{spectra2structure2024} & UV & $0.9$ & --- & $7.2$ & $0.158$ & $98.5$ & --- & --- & --- & --- & --- \\
    \quad SMEN--SD~\citep{spectra2structure2024} & IR & $60.1$ & --- & $75.9$ & $0.823$ & $97.9$ & --- & --- & --- & --- & --- \\
    \quad SMEN--SD~\citep{spectra2structure2024} & Raman & $65.6$ & --- & $77.9$ & $0.869$ & $97.8$ & --- & --- & --- & --- & --- \\
    \quad SpectraLLM~\citep{spectrallm2026} & UV & $0.0$ & --- & $2.8$ & $0.098$ & \best{$99.9$} & --- & --- & --- & --- & --- \\
    \quad SpectraLLM~\citep{spectrallm2026} & IR & $0.5$ & --- & $16.1$ & $0.200$ & $99.1$ & --- & --- & --- & --- & --- \\
    \quad SpectraLLM~\citep{spectrallm2026} & Raman & $0.8$ & --- & $13.4$ & $0.197$ & $98.9$ & --- & --- & --- & --- & --- \\
    \quad SpectraLLM~\citep{spectrallm2026} & All spec. & $0.5$ & --- & $19.2$ & $0.208$ & $98.9$ & --- & --- & --- & --- & --- \\

    \midrule
    \multicolumn{12}{@{}l}{\textit{Diffusion spectra-to-3D}} \\[2pt]
  
    \quad DiffSpectra~\citep{diffspectra2025}& UV$^{\ddagger}$    & $0.0$         & $1.762$ & $0.0$  & $0.220$ & $0.0$  & --- & --- & --- & $1{,}000$ & $475$ \\
    \quad  & IR$^{\ddagger}$    & $0.0$         & $1.578$ & $0.2$  & $0.201$ & $1.0$  & --- & --- & --- & $1{,}000$ & $473$ \\
    \quad  & Raman$^{\ddagger}$ & $0.8$         & $1.560$ & $6.6$  & $0.233$ & $46.0$ & --- & --- & --- & $1{,}000$ & $472$ \\
    \addlinespace[2pt]
    \quad             & All spec. & $60.3$    & $1.539$ & $96.1$ & $0.831$ & $97.9$ & --- & --- & --- & $1{,}000$ & $469$ \\
    \midrule
    \multicolumn{12}{@{}l}{\textit{Supervised 3D-to-spectra}} \\[2pt]
    \quad DetaNet~\citep{zou2023qm9s} & 3D & ---         & ---          & ---          & ---          & ---          & $0.656$      & $0.858$      & $0.818$      & --- & --- \\

    \midrule
    \multicolumn{12}{@{}l}{\textit{Multimodal any-to-any (Ours)}} \\[2pt]
    \quad \best{MOSAIC}           & UV    & $0.0$       & $1.407$      & $15.1$      & $0.149$      & $77.3$      & ---          & 0.627          & 0.590          & $50$ & $315$ \\
    \quad            & IR    & $73.5$      & $0.433$      & $93.8$      & $0.870$      & $93.0$      & 0.918          & ---          & 0.979          & $50$ & $315$ \\
    \quad          & Raman & $79.6$      & $0.357$      & $94.7$      & $0.907$      & $93.7$      & 0.926          & 0.972          & ---          & $50$ & $315$ \\
    \quad      & 3D & ---         & ---          & ---          & ---          & ---          & \best{$0.856$} & \best{$0.868$} & \best{$0.852$} & $50$ & $1{,}172$ \\
    \quad        & All spec. & \best{$88.8$} & \best{$0.265$} & \best{$98.6$} & \best{$0.949$} & $96.5$ & ---          & ---          & ---          & $50$ & $315$ \\
    \bottomrule
  \end{tabular}
  }
\end{table}

We benchmark MOSAIC against published baselines and evaluate
structure-to-spectrum generation for retrieval against
experimental spectra (Section~\ref{sec:exp:struct2ir}).
We then ablate key model components and examine how performance
scales with model size and training set size
(Section~\ref{sec:exp:scaling}).

\paragraph{Evaluation metrics.}
We evaluate structure and spectra generation via molecular graph recovery, geometric accuracy,
chemical validity, and spectral similarity.
\textbf{Acc@1} measures the fraction of test molecules whose
Top-1 prediction matches the ground-truth molecular graph,
including atom and bond types. We evaluate one generated
sample for diffusion and flow matching models, and the highest-ranked
candidate for autoregressive baselines.
\textbf{H-RMSD} measures coordinate error in \AA{} after Kabsch
alignment followed by Hungarian atom assignment based on
coordinate distances, without using atom types.
\textbf{Comp} is the fraction of predictions with the correct
molecular formula.
\textbf{Tani\textsubscript{MG}} measures Tanimoto similarity
between Morgan fingerprints.
\textbf{Valid} is the fraction of generated molecular graphs
that pass RDKit sanitization.
\textbf{Spec cos} measures cosine similarity between generated
and reference spectra in the log-transformed, standardized
spectral space, reported separately for UV, IR, and Raman
in Table~\ref{tab:headline}.

\paragraph{Dataset.}
We train and evaluate MOSAIC (and baselines) on QM9S~\citep{zou2023qm9s},
which pairs molecular structures from QM9~\citep{qm9_2014}
with simulated UV, IR, and Raman spectra.
Our split contains $110{,}000$ training molecules,
$10{,}000$ validation molecules, and $8{,}057$ test molecules.
We evaluate on the full test split, with training and evaluation
details provided in Appendices~\ref{sec:app:phase2}
and~\ref{sec:app:eval}.

\paragraph{Baselines.} We benchmark MOSAIC against three autoregressive spectra-to-SMILES baselines: Alberts et al.~\citep{alberts2024irstructure}, SMEN--SD~\citep{spectra2structure2024}, and SpectraLLM~\citep{spectrallm2026}. Implementation details are provided in Appendix~\ref{sec:app:baselines}. We also compare against DiffSpectra~\citep{diffspectra2025}, a diffusion-based spectra-to-3D structure model, and its single-modality ablations. We retrain DiffSpectra on the full canonical QM9S training split of $110$k molecules. The model of Alberts et al. receives the ground-truth molecular
formula as additional conditioning information, which is not
provided to MOSAIC, and uses beam search with a beam width
of $10$ at inference. The three autoregressive baselines generate neither 3D coordinates nor spectra, so H-RMSD and spectrum-generation metrics are not applicable. Both MOSAIC and DiffSpectra receive the ground-truth atom count
for spectra-to-structure generation in
Table~\ref{tab:headline}. We also evaluate a separate MOSAIC
variant that infers the atom count from spectra, achieving
$85.5\%$ Acc@1 
(Section~\ref{sec:exp:variable_count}).

\subsection{Results}
\label{sec:exp:headline}

\paragraph{Spectra-to-structure generation.} We compare MOSAIC against the baseline models for the key spectra-to-structure task in Table~\ref{tab:headline}. We present two settings 1) individual spectra as inputs, 2) all 3 spectra as inputs. In the latter, MOSAIC improves structure generation over DiffSpectra by $28.5$\% ($88.8\%$ vs. $60.3\%$), while reducing H-RMSD from $1.539$ to $0.265$. Concomitantly, MOSAIC also achieves the best Tani\textsubscript{MG} and Comp across all models.

\paragraph{Any-to-any generation.}
A single trained MOSAIC model supports generation in both
directions between molecular structures and spectra.
When conditioned on molecular structures, it generates UV,
IR, and Raman spectra with higher cosine similarity to
ground truth than DetaNet, a specialized structure-to-spectra
predictor (Table~\ref{tab:headline}).
For structure generation on QM9S, MOSAIC achieves Acc@1 of 73.5\% and 79.6\% using IR and Raman, respectively, as the sole spectral input.
Unlike the specialized baselines, MOSAIC supports both generation directions and different combinations of spectral inputs within a single model. This flexibility is especially relevant to laboratory workflows, where only one or two spectral modalities may be available. Performance nevertheless depends on the conditioning modality. UV-only conditioning yields $0\%$ Acc@1, despite high cosine similarity for structure-to-UV generation. We hypothesize that this asymmetry reflects greater ambiguity in recovering exact molecular connectivity from UV electronic transitions than from IR and Raman vibrational fingerprints.

\paragraph{Structure generation without atom count known a priori.}
\label{sec:exp:variable_count}
We also evaluate a MOSAIC variant that jointly infers molecular
size and structure from spectra, without receiving the ground-truth
atom count. On the full QM9S test set, this variant achieves
$85.5\%$ Acc@1, compared with $88.8\%$
for count-conditioned MOSAIC (Table~\ref{tab:headline}).
It predicts the atom count and molecular formula correctly for $95.8\%$
and $95.1\%$ of test molecules, respectively. H-RMSD is
$0.233$~\AA{} when evaluated on ground-truth occupied slots.
Evaluation details are provided in Appendix~\ref{sec:app:eval}. These results show that MOSAIC can generate molecular structures without requiring the atom count as an input.

\paragraph{Matching generated spectra to experimental spectra.}
\label{sec:exp:struct2ir}
We evaluate whether structure-conditioned spectrum generation can
retrieve a molecule's experimental IR spectrum from a fixed database
of $99$ gas-phase IR spectra~\citep{eurochampIR}.
For each molecule, we use MOSAIC to generate an IR spectrum from its structure and
rank all experimental spectra by cosine similarity to the generated
spectrum. We compare MOSAIC against an adapted DetaNet direct-spectrum
predictor using the same candidate database.
As shown in Table~\ref{tab:struct2ir}, MOSAIC ranks the correct spectrum
first for $15.2\%$ of molecules, compared with $5.1\%$ for DetaNet,
and among the top five for $30.3\%$, compared with $17.2\%$.
Mean matched-spectrum cosine similarity is $0.308$ for MOSAIC versus
$0.194$ for DetaNet. These results indicate simulation-to-experiment transfer, 
where MOSAIC outperforms baselines in retrieving experimental IR spectra (including spectra with contaminants) despite being trained solely on simulated QM9S data. Evaluation details and limitations are provided in
Appendix~\ref{sec:app:struct2ir}.

\subsection{Ablations}
\label{sec:exp:scaling}

\begin{figure}[t]
  \centering
  \includegraphics[width=\linewidth]{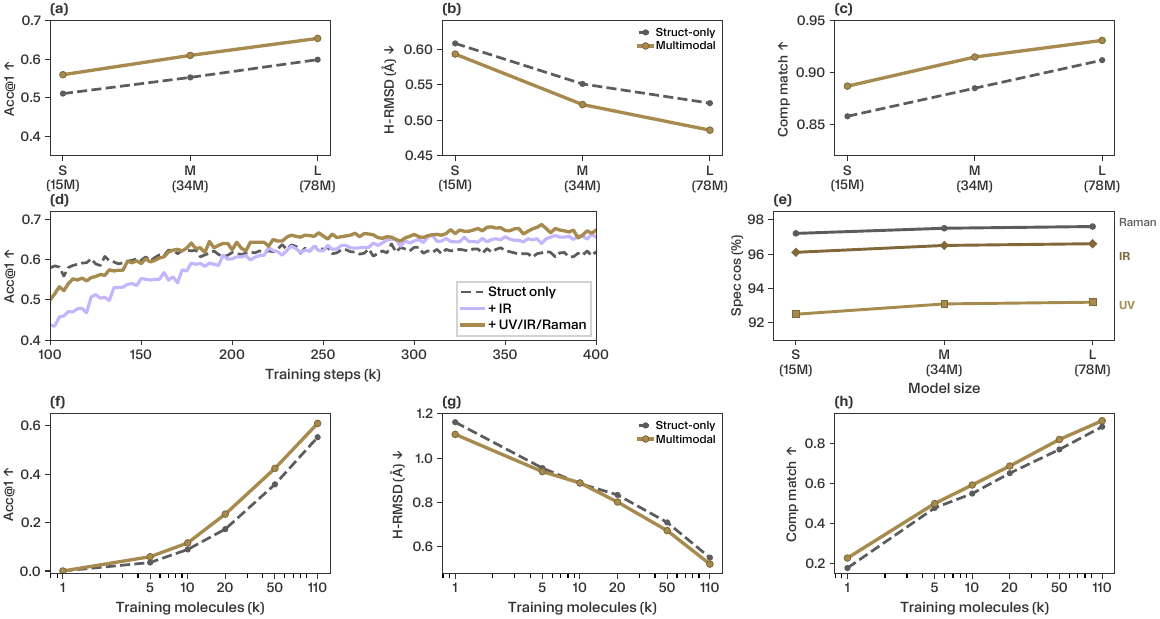}
  \caption{\textbf{Model and data scaling.} \textbf{(a--c)}~Acc@1, H-RMSD and Composition across model scale. \textbf{(d)}~The structure-only model overfits, while the multimodal model (gold) remains stable. Adding IR only as a co-target (purple) outperforms struct-only and captures most of the 3-spectra advantage. \textbf{(e)}~Spectra generation quality across model scale. \textbf{(f--h)}~Data scaling with an M-scale decoder. }
  \label{fig:scaling}
\end{figure}

\paragraph{Multimodal encoder pretraining.}
We ablate the three encoder pretraining objectives: masked
reconstruction, cross-modal contrastive alignment, and EMA
self-distillation. Masked reconstruction contributes most
to encoder performance, followed by contrastive alignment
and self-distillation. Detailed results are provided in
Appendix~\ref{sec:app:phase1_ablation}.

\paragraph{Multimodal any-to-any flow matching training.}
We study the influence of pretraining the multimodal encoder before generative training. We compare three settings, frozen pretrained, pretrain-then-finetune end-to-end and randomly initialized then trained end-to-end (Table \ref{tab:attribution}). The pretrain-then-finetune approach improves structure generation over frozen pretraining and random initialization by approximately $25$ and $13$\%, respectively. More details are available in Appendix \ref{sec:app:attribution}.

Next, we ablate the contribution of spectral generation
targets by training a structure-only variant,
MOSAIC-struct-only. During flow matching training, we set
$\pi_{\mathrm{struct}}=1$ and
$\pi_{\mathrm{UV}}=\pi_{\mathrm{IR}}=\pi_{\mathrm{Raman}}=0$,
so that spectra serve only as conditioning inputs.
This ablation tests whether jointly learning to generate
spectra improves structure generation.
Figure~\ref{fig:scaling} compares the two models across
model sizes and training-set sizes.

\paragraph{Model scaling.}
We compare decoder sizes from S ($15$M) to L ($78$M), training
each on $110$k molecules for $400$k steps
(Figure~\ref{fig:scaling}a--c). We select each model's checkpoint independently using the same validation criterion. Both MOSAIC and MOSAIC-struct-only
improve with model size, with MOSAIC achieving roughly 5\% higher Acc@1 across S, M, and L.
Training curves reveal a further
difference (Figure~\ref{fig:scaling}d): structure generation for
MOSAIC-struct-only peaks around $280$--$330$k steps before
declining. Adding IR as a prediction target reduces this decline,
while joint training on structure and all three spectra achieves
the strongest and most stable performance.
We observe similar trends across model sizes
(Figure~\ref{fig:convergence}), suggesting that learning
to predict spectra provides complementary supervision
that improves structure generation.
Structure-to-spectra generation also improves
with model size (Figure~\ref{fig:scaling}e), showing that scaling
benefits both generation directions.

\paragraph{Data scaling.}
MOSAIC achieves better structure generation than MOSAIC-struct-only
across all training-set sizes (Figure~\ref{fig:scaling}f).
The advantage grows from $2.7$\% at $10$k
training molecules to $5.7$\% at $110$k, with similar
trends in H-RMSD and molecular formula prediction
(Figure~\ref{fig:scaling}g--h). These results suggest that the
benefit of multimodal training extends beyond low-data settings
(Table~\ref{tab:scaling}).

\begin{figure}[tp]
  \centering
  \includegraphics[width=\linewidth]{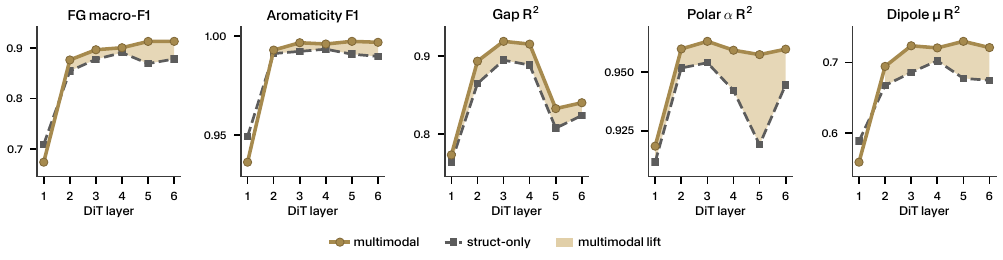}
  \caption{\textbf{Layer-wise probes on MOSAIC structure tokens.} From DiT layer 2 onward, the multimodal MOSAIC representations are richer than struct-only representations for every prediction task. Metrics are averaged over noise levels. See Appendix~\ref{sec:app:probing} for the encoding and probing protocol.}
  \label{fig:probe}
\end{figure}

\paragraph{Learned representations.}
We assess the learned structure representations by freezing each generator
and training probes across six layers to predict five molecular properties.
Figure~\ref{fig:probe} shows that MOSAIC achieves higher noise-averaged scores
than MOSAIC-struct-only on all five tasks from layer two onward.
These results are consistent with spectral co-targets providing chemically
meaningful auxiliary supervision, helping the model encode information
about related molecular properties in its structure-token representations.

\section{Conclusion}
\label{sec:conclusion}

We introduce MOSAIC, a multimodal flow matching model that
combines a self-supervised encoder with an MMDiT generator
for any-to-any generation across molecular structures and
UV, IR, and Raman spectra. On QM9S, MOSAIC outperforms the
strong specialist spectra-to-structure and structure-to-spectra baselines, while enabling reverse generation and missing-modality completion using
a \textit{single} model. Ablations and scaling studies demonstrate
the benefits of multimodal any-to-any training for structure and spectral generation. Linear probes further suggest that multimodal training
yields richer molecular representations than structure-only
training. We further demonstrate the potential utility of bidirectional generation for experimental settings by linking molecular structures to experimental IR spectra.

\newpage
\begin{ack}
This work was funded by Lila Sciences, where all authors are employed. The authors declare no other competing interests.
\end{ack}

\bibliographystyle{unsrtnat}
\bibliography{refs}

\clearpage
\appendix

\clearpage

\section*{Author Contributions}
T.P. conceived the study, developed and implemented the MOSAIC framework, conducted experiments and analyses, and wrote the manuscript. E.P. synthesized analysis of the MOSAIC framework, implemented and evaluated baselines, contributed to manuscript writing and prepared figures. R.G.-B. provided scientific guidance and research direction, and reviewed the manuscript. S.M. supervised the research, coordinated the project, and reviewed the manuscript.

\section*{License}
This work is licensed under
\href{https://creativecommons.org/licenses/by-nc-nd/4.0/}
{CC BY-NC-ND 4.0}.

\section{Limitations and Broader Impacts}
\label{sec:limitations}

All structure generation results are on QM9-sized molecules ($\leq\!9$ heavy atoms) with simulated DFT spectra. Spectra-to-structure generation under experimental noise, larger conformational diversity, and drug-like molecular sizes remains untested. The scaling results describe improvements within QM9S, whose training and test sets cover similar chemical space. As with any molecular structure generator, deployment at larger scales could pose misuse risks in unsafe chemistry applications, and would require domain-specific safeguards, screening, and human oversight before practical use.

The standard structure-generation benchmark uses a supplied atom-count/active-slot mask. The variable-count extension removes this inference-time input, but retains a fixed maximum slot budget and is evaluated here on the same test split (Appendix~\ref{sec:app:eval}). 

\section{Implementation details}
\label{sec:app:impl}

This section provides full hyperparameters for the MOSAIC model, the UAE-3D structure autoencoder (Appendix \ref{sec:app:uae}), the spectral tokenizer (Appendix \ref{sec:app:patches}), the self-supervised multimodal encoder (Appendix \ref{sec:app:phase1}), and the multimodal flow matching generator (Appendix \ref{sec:app:phase2}).

\subsection{UAE-3D structure autoencoder}
\label{sec:app:uae}

The structure tokenizer is a UAE-3D-style variational autoencoder~\citep{uae3d2025} over atom-coordinate pairs. We train it from scratch on the same $110$k-molecule QM9S training set as encoder pretraining and generative training (Appendix~\ref{sec:app:dataset}). The trained UAE is frozen during downstream training. The variable count extension uses a separate UAE checkpoint whose atom decoder predicts six classes: empty, H, C, N, O, and F (Appendix~\ref{sec:app:phase2}).

\textbf{Architecture.} The standard UAE encoder comprises $6$
relational-attention blocks of width $64$ with $8$ heads. Its node
inputs concatenate $44$-dimensional RDKit atom features with $3$D
coordinates. Edge features comprise bond-type indicators, a self-loop
indicator, and a Gaussian-basis expansion of squared interatomic
distances. The encoder predicts a $16$-dimensional posterior mean and
log-variance for each atom. The decoder comprises $4$ pre-norm
transformer blocks of width $64$ with $8$ heads and predicts per-atom
coordinates, atom-type logits over $\{\mathrm{H,C,N,O,F}\}$, and
pairwise bond-type logits. Input coordinates are centred and scaled per molecule. After training, posterior means are standardized using
a single scalar mean and standard deviation computed across all latents in the training set before use by the
generator.

\subsection{Spectral patching and normalization}
\label{sec:app:patches}

UV, IR, and Raman spectra are tokenized using linear projections of fixed-length
one-dimensional patches, analogous to image patch embeddings. We use $701$ UV samples
on a $1$--$15$~eV grid ($0.02$~eV spacing), and $3{,}501$ samples each
for IR and Raman over $500$--$4{,}000$~cm$^{-1}$ ($1$~cm$^{-1}$ spacing). For the encoder, we use overlapping patches
of length $20/50/50$ with stride $10/25/25$ for UV/IR/Raman, producing
$69 + 139 + 139 = 347$ tokens per molecule. For the generator, we use non-overlapping
patches of length $10/25/25$ with the same strides, producing
$70 + 140 + 140 = 350$ tokens. We normalize all spectra using a two-step procedure. First, we apply
$x \mapsto \log(1+x)$ to compress the long tail of high-intensity features.
Second, we standardize each log-transformed spectrum independently across its spectral bins by subtracting its mean and dividing by its standard deviation.

\subsection{Multimodal encoder pretraining}
\label{sec:app:phase1}

The self-supervised multimodal encoder described in Section~\ref{sec:method:phase1}
is an $8$-block transformer with width $512$ and $8$ attention heads. It is
trained on the $110$k-molecule training set for $100$k steps using AdamW, with
learning rate $3\!\times\!10^{-4}$, minimum learning rate $10^{-5}$, weight decay
$0.05$, maximum gradient norm $1.0$, a $2{,}000$-step warmup, and batch size
$128$. Training was performed on a single H100 GPU and took approximately
$12$ GPU-hours. For masked input construction, we apply modality dropout with probability $0.2$,
followed by token masking with probability $0.4$ within each remaining modality.
Each modality has a linear reconstruction head applied directly to
the encoder's output states. We minimize MSE against normalized spectral patches
or UAE-3D structure latents at masked positions, including those in dropped
modalities. Padded atom slots are excluded from this loss. The structure
reconstruction term is weighted by a factor of $3$.

We also include a cross-modal contrastive loss and a self-distillation loss. First, we use a symmetric
InfoNCE objective \citep{infonce2018} with temperature $0.07$, averaged over all
$\binom{4}{2}=6$ pairs of per-modality CLS tokens. Second, we match mean-pooled student states from block $3$ to
EMA-teacher states from block $6$. Pooling covers the full token sequence,
including CLS tokens, padded atom slots, and dropped-modality positions. The
teacher sees the same modality dropout as the student, but no token masking.
Each pooled vector passes through a two-layer MLP with $128$ output dimensions,
using an EMA copy of the student. After L2 normalization,
we minimize $2-2\cos(s,t)$ between the projected student and teacher vectors.
The teacher EMA decay is $0.9999$. The final encoder pretraining loss uses weights
$\lambda_\mathrm{recon}=1.0$, $\lambda_\mathrm{contr}=0.5$, and
$\lambda_\mathrm{distill}=0.5$.

For generator conditioning, the encoder keeps a fixed $380$-slot sequence:
$69$ UV patches, $139$ IR patches, $139$ Raman patches, $29$ atom slots,
and four CLS tokens. Spectral patches contain $20$, $50$, and $50$ grid points,
with strides of $10$, $25$, and $25$, respectively. A withheld modality retains
its slots, but its input content is replaced by learned mask embeddings together
with the position, modality, and CLS embeddings. Its output states therefore
reflect the observed context without exposing the withheld input. The full
encoder output supplies the generator's attention context. The four CLS states are
concatenated in UV, IR, Raman, and structure order and linearly projected to a
global adaLN conditioning vector, which controls feature scales, shifts, and
residual gates in every MMDiT block. For the main benchmark, this encoder is finetuned end-to-end together
with the DiT generator described in Section~\ref{sec:app:phase2}, rather than kept
frozen. The frozen-encoder setting is reported separately in the $3\!\times\!2$
attribution table in Appendix~\ref{sec:app:attribution}.

\subsection{Multimodal flow matching training}
\label{sec:app:phase2}

The MOSAIC generator is a $298$M-parameter MMDiT. Training uses a $300$k-step budget on
$4 \times$~H100 GPUs with batch size $256$ ($64$ per GPU). Results
in Table~\ref{tab:headline} are reported from the EMA checkpoint. We describe the architecture, optimization,
target sampling, timestep schedule, and classifier-free guidance below. Within each matched experimental setting in Section~\ref{sec:exp:scaling},
the struct-only ablation uses the same architecture and training hyperparameters
as its multimodal counterpart, changing only the target sampler:
$\pi_\textrm{struct}=1$, with all spectral target probabilities set to zero.

\paragraph{Architecture.}
The MMDiT contains $D=14$ blocks with width $768$ and $12$ attention heads. Each block keeps separate QKV projections, adaLN
modulation, and MLPs for the noisy target stream and the encoder context stream.
The two streams are then concatenated and processed with joint scaled dot-product
attention over the concatenated $\text{target}\,\Vert\,\text{context}$ sequence, using
RMSNorm Q/K-normalization \citep{sd32024}.

The target stream consists of noisy spectral patch tokens and noisy UAE-3D
structure latents. Target tokens receive modality embeddings, with sinusoidal
position embeddings for spectra and learnt per-atom position embeddings for
structure. The context stream contains the clean per-modality CLS tokens and
per-patch hidden states produced by the multimodal encoder, projected from
$d_\mathrm{enc}=512$ to the MMDiT width $d=768$. The encoder's CLS pools are
concatenated and mapped to a single $768$-dimensional adaLN conditioning vector,
which is broadcasted over the target stream. Each target token's adaLN conditioning
is the sum of its modality-specific timestep embedding and the projected global CLS vector. The output
uses separate velocity heads for each modality, projecting back to patch length
$10/25/25$ for UV/IR/Raman and to the UAE-3D latent dimension $16$ for
structure. During structure denoising, we deliberately do \emph{not} provide clean atom-type embeddings on the target stream. 

\paragraph{Loss.}
The main training objective is a per-modality rectified-flow
\citep{rectifiedflow2022} velocity MSE. We use modality weights
$w_\textrm{struct}=1.5$, $w_\textrm{UV}=0.40$, and
$w_\textrm{IR}=w_\textrm{Raman}=0.20$. In addition, we add an auxiliary
atom-type cross-entropy loss on the predicted structure with weight
$\lambda_\mathrm{ce}=0.1$.

For each modality, Equation~\ref{eq:p2} uses the masked mean

\begin{equation}
  \ell_m = \frac{\sum_{b,j} a_{bj}^m\,
    \|v_{\theta,bj}^m-(\varepsilon_{bj}^m-x_{0,bj}^m)\|_2^2}
    {d_m\,\max(1,\sum_{b,j}a_{bj}^m)}.
\label{eq:masked-flow-loss}
\end{equation}

Here $a_{bj}^m=1$ for valid tokens $j$ in examples $b$ targeting modality $m$, and $0$ otherwise; $d_m$ is the token feature dimension. Thus $\ell_m=0$ when no tokens are selected. These reductions use each worker's minibatch, with gradients averaged across workers.

To compute the atom-type cross-entropy term, we reconstruct
$\hat{x}_0 = x_t - t \cdot v_\theta$ from the predicted velocity, pass
the unstandardized $\hat{x}_0$ through the frozen UAE-3D decoder, and apply cross-entropy to the
resulting atom-type logits against the true atomic numbers. This term is applied
as a mean over valid atoms in structure-target examples and is zero when no such atoms are selected.

\paragraph{Target sampling.}
During multimodal training, we draw independent Bernoulli target indicators per
example with $\pi_\textrm{struct}=0.6$ and
$\pi_\textrm{UV}=\pi_\textrm{IR}=\pi_\textrm{Raman}=0.2$.
If no modality is selected, we select structure as the target. These
probabilities apply before this fallback.

\paragraph{Per-modality timestep schedule.}
Each target modality receives its own timestep,
$t^m=\sigma(\mu_m+u_m)$, where $u_m\sim\mathcal{N}(0,1)$ independently and $\sigma$ is the sigmoid. We set
$\mu_\textrm{struct}=0$ for structure and
$\mu_\textrm{UV}=\mu_\textrm{IR}=\mu_\textrm{Raman}=2$ for spectra, which biases
the spectral targets toward higher-noise training examples. The choices of
structure and spectral $\mu$ are ablated in Appendix~\ref{sec:app:timestep_mu}.

\paragraph{Classifier-free guidance.}
During training, the encoder context tokens and CLS states are jointly masked per example with probability $p_\textrm{drop}=0.1$. At inference, we use
classifier-free guidance scale $w_\textrm{cfg}=3$ unless stated otherwise. The
structure-conditioned IR retrieval study in Appendix~\ref{sec:app:struct2ir} uses
$w_\textrm{cfg}=3$ and reports both per-molecule cosine similarity
and rank-based metrics.

\paragraph{Optimization.}
We train with AdamW using peak learning rate $3\!\times\!10^{-4}$, minimum
learning rate $3\!\times\!10^{-5}$ and weight decay $0.05$.
We clip only the DiT gradients to a maximum norm of $1.0$. The schedule uses a $2{,}000$-step linear
warmup followed by cosine decay over the training horizon. Training is
performed in bf16 mixed precision with PyTorch DDP on
$4\!\times\!$H100, using global batch size $256$. Random
structure rotations are applied to every training sample, while spectra are not
augmented. We maintain an EMA copy of the DiT with decay $0.9999$ and use this
copy for all reported results.

\paragraph{Variable atom count model and training.}
The variable-count model uses $29$ structure slots, matching the maximum number of atoms in QM9 when hydrogens are included. Each slot has a $16$-dimensional latent representation and can decode as an atom or empty. A separately trained, frozen UAE predicts six atom classes: empty, H, C, N, O, and F. The generator has $14$ blocks of width $768$ with $12$ attention heads. Its width-$512$ conditioning encoder is initialized randomly and trained jointly with the generator.

We use a training batch size of $256$ and EMA decay $0.9999$.
After $2{,}000$ warmup steps, the learning rate decays from
$10^{-4}$ to $10^{-5}$. For this variant, the timestep distribution uses $\mu_m=0$ for all four modalities. Flow loss weights for structure, UV, IR, and Raman are $1$, $0.4$, $0.2$, and $0.2$, respectively. Decoded and direct-head atom and bond cross-entropies each have weight $0.1$ in the final objective. During training, atom classification supervises both occupied and empty slots, while the latent flow matching loss is computed only on slots occupied in the ground-truth molecule. At inference, the model predicts slot occupancy without receiving the ground-truth occupancy mask or atom count.

\subsection{Inference and evaluation}
\label{sec:app:eval}

For the standard QM9S benchmark, metrics use the $8{,}057$-molecule test set.
Sampling uses $50$ Euler ODE steps, a polynomial timestep schedule
$t_i=((N-i)/N)^p$ with $N=50$, $i=0,\ldots,N$, and $p=2$, CFG scale $w_\textrm{cfg}=3$, and EMA DiT weights.
With CFG scale $3$, the $50$ Euler steps require $100$ network evaluations.
Spectra-to-structure timings use batches of $64$ on one H100 and exclude final UAE decoding.
The polynomial schedule allocates more steps near $t\to 0$. For paired
comparisons, such as struct-only versus multimodal generation at a fixed
checkpoint, we reuse the same initial Gaussian noise for each molecule.

Generated structures are decoded from the predicted UAE-3D latent using the
frozen autoencoder. We then evaluate them with RDKit sanitization,
Hungarian-aligned RMSD against the ground-truth coordinates, and exact atom- and
bond-labelled graph isomorphism, reported as Acc@1. Tanimoto similarity is computed
using Morgan fingerprints with radius~2 and 2048~bits. Tanimoto similarity is averaged over prediction--reference pairs
for which fingerprint similarity is successfully computed;
unscored pairs are excluded. For MOSAIC, both molecules must
additionally pass RDKit sanitisation. Unsuccessful pairs are excluded.
Spectral quality is measured by cosine
similarity, computed in the same log-then-$z$-score
normalized spectral domain used throughout preprocessing
(Appendix \ref{sec:app:patches}).

RMSD\textsubscript{DS} follows DiffSpectra's two-stage Hungarian
assignment, with Kabsch alignment between stages. Assignment costs
combine Euclidean distance with atom-type penalties: $0$ for
identical types, $2$ for substitutions among C, N, and O, and $10$
otherwise. Pairs with costs above $5$ are discarded. Coordinate
RMSD is averaged over successfully matched molecules with at least
three retained pairs. Both RMSD\textsubscript{DS} and H-RMSD are
reported in \AA{}.

\paragraph{Variable atom count inference and evaluation.}
We evaluate the model trained for $1$M steps on all $8{,}057$
test molecules, using UV, IR, and Raman spectra as conditioning.
We select the checkpoint with the highest validation graph accuracy
under EMA weights and use these weights for sampling, with $50$
Euler steps on a polynomial schedule of power $2$ and CFG $=1$.

The model generates structures over $29$ slots without access to
the ground-truth atom count or occupancy mask. The UAE decoder
predicts an atom type or an empty class for each slot, and we
retain only the predicted atoms when computing Acc@1 and
composition match. Atom-count accuracy measures how often the
predicted total number of atoms, including hydrogen, is correct.
The mean absolute error in atom count is $0.0581$. We evaluate H-RMSD on the ground-truth occupied slots using the
alignment and matching procedure described above. This isolates
coordinate quality from errors in the predicted atom count,
so missing or extra decoded atoms are not penalized by this metric.
The illustrative trajectory in Figure~\ref{fig:crossmodal} uses
a separate checkpoint with the same $50$ sampling steps and CFG $=1$.

\begin{figure}[h!]
  \centering
  \includegraphics[width=\linewidth]{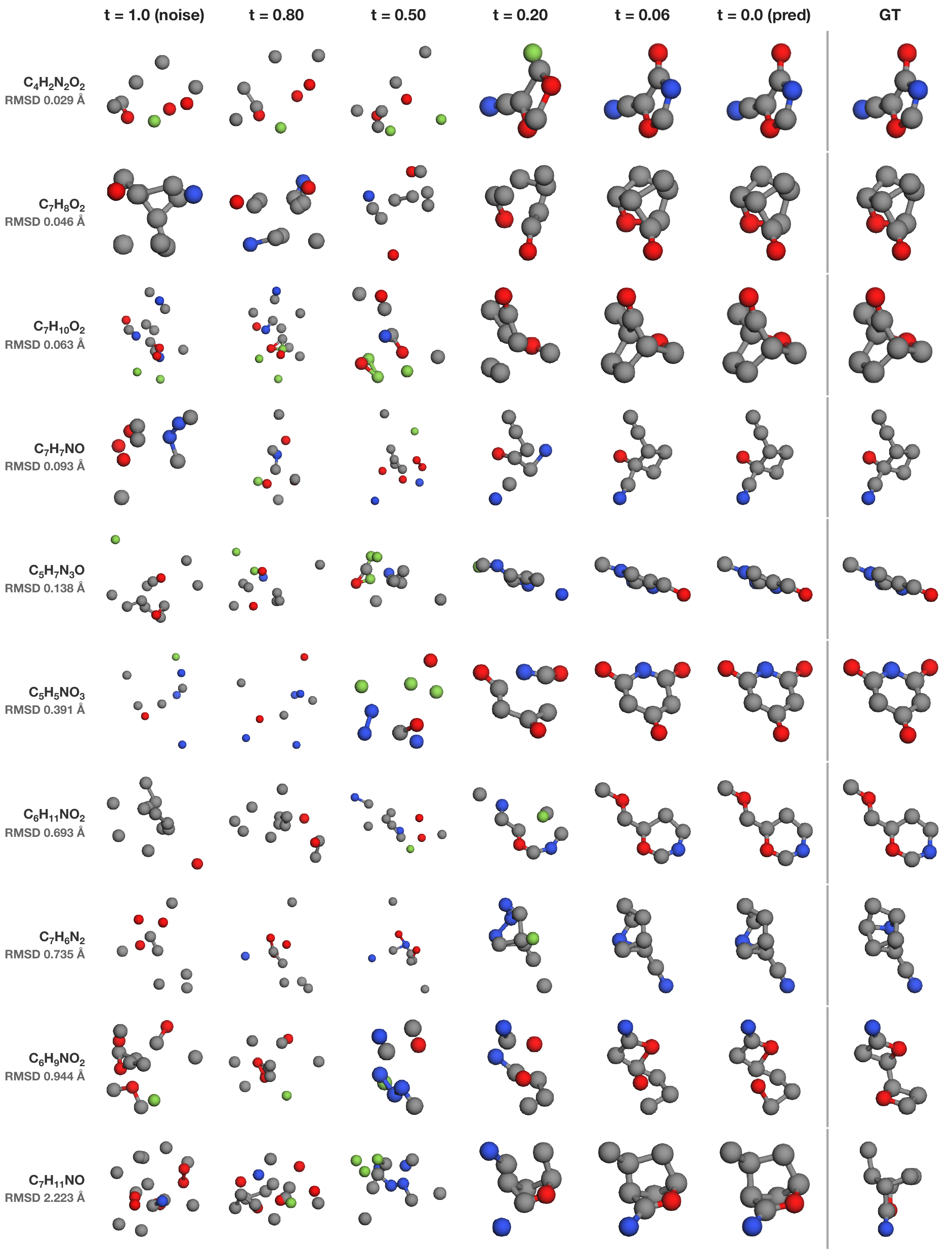}
  \caption{\textbf{Denoising trajectories for molecular structure generation.} Ten molecules are ordered by increasing final H-RMSD. Six snapshots show the progression from noise ($t=1$) to the predicted structure ($t=0$) over 50 Euler steps. Ground-truth structures appear in the rightmost column. Each structure is independently scaled to fit its panel. Heavy atoms are colored by element (C: dark gray, N: blue, O: red, F: green), with hydrogens omitted for clarity.}
  \label{fig:trajectory}
\end{figure}

\subsection{Layer-wise probes}
\label{sec:app:probing}

We freeze each generator and encode clean structures with the frozen UAE to obtain normalized latents $z_0$. We construct $z_t=(1-t)z_0+t\epsilon$, where $\epsilon\sim\mathcal{N}(0,I)$ and $t\in\{0,0.25,0.5,0.75\}$, and supply these structure target tokens with matching timesteps. Both models receive clean UV, IR, and Raman inputs through the conditioning encoder. The multimodal model additionally receives spectral target tokens corrupted at the same noise level; the structure-target-only model receives no spectral target tokens. We mean-pool valid structure tokens at each probed layer, fit a separate probe for each layer and noise level, and average the resulting scores over noise levels. These probes compare conditioned representations under the two target-token configurations.

\clearpage
\section{Baselines}
\label{sec:app:baselines}

All four spectra-to-structure baselines in Table~\ref{tab:headline} are evaluated on the same QM9S reference split as MOSAIC.
The computation of Acc@1 depends on each model's output format.
For MOSAIC and DiffSpectra, we apply the VF2 algorithm to test exact graph isomorphism, matching atom and bond labels with explicit hydrogens included. These graphs are constructed directly from the predicted and reference labels without conversion to SMILES.
For SMEN--SD and SpectraLLM, we parse the predicted and reference SMILES with RDKit, remove explicit hydrogens, and apply VF2 to the resulting graphs, matching atomic numbers and bond types.
For Alberts et al., we take the highest-ranked beam prediction and require an exact match between the predicted and reference canonical isomeric SMILES after hydrogen removal. H-RMSD requires atomic coordinates, so we report it only for DiffSpectra and MOSAIC, which both predict 3D structures.

\paragraph{Alberts et al.~\citep{alberts2024irstructure}}
Alberts et al.\ train a transformer to translate a tokenized molecular-formula prefix together with $400$ binned IR intensities into a canonical SMILES string, conditioning on the published merged spectral window. We retrain the published model from scratch on our QM9S split, using the published architecture, source-token format, and training procedure. For inference, we use the published beam-search ranking.

\paragraph{SMEN--SD~\citep{spectra2structure2024}.}
SMEN--SD combines an EGNN molecular encoder, a ViT spectral encoder,
and a transformer SMILES decoder. The encoders are trained with a
CLIP-style contrastive objective, while the decoder learns to predict
SMILES from the spectral representation. Training uses randomized SMILES
targets; generated strings are canonicalized during evaluation.
We use a training budget of
$1{,}000$ epochs, a batch size of $400$, and a learning rate of $10^{-4}$.

\paragraph{SpectraLLM~\citep{spectrallm2026}}
SpectraLLM fine-tunes Qwen3-32B with LoRA to predict canonical SMILES strings from a natural language prompt that lists each spectrum's detected peaks (wavenumbers, intensities, and widths). We adapt the released two-stage training setup to our QM9S split. Stage~1 pretrains a single LoRA adapter with rank $8$ across all linear layers, using a learning rate of $5\!\times\!10^{-5}$ for one epoch on the all-spectra training set. Stage~2 separately fine-tunes modality-specific adapters from the shared Stage 1 checkpoint.

\paragraph{DiffSpectra~\citep{diffspectra2025}}
DiffSpectra combines a SpecFormer spectral encoder with a Diffusion
Molecule Transformer decoder, trained to jointly denoise atom features,
bond features, and 3D coordinates conditioned on multiple spectra.
We retrain the model on our QM9S training split with a training budget
of $1$ million iterations on two GPUs, initializing SpecFormer from
a separate pretraining stage on the training molecules.
Inference uses $1{,}000$ DDPM sampling steps. The single-modality
DiffSpectra rows (UV, IR, Raman) evaluate the same all-spectra model
with the unused spectral inputs set to zero.

\paragraph{DetaNet~\citep{zou2023qm9s}}
DetaNet is an E(3)-equivariant tensor-attention network
introduced alongside QM9S. Its original spectral prediction
pipeline reconstructs IR and Raman spectra from predicted
Hessian matrices and derivatives of dipole moments and
polarizabilities, while predicting UV--Vis spectra directly.
We adapt DetaNet to jointly predict UV, IR, and Raman spectra,
using a shared encoder with three message-passing blocks
($N=3$), 256 feature channels ($F=256$), and an interatomic
distance cutoff of $R_c=5.0$~\AA. We attach a separate MLP
prediction head for each spectral modality.

\clearpage
\section{Scaling study}
\label{sec:app:scaling}
\label{sec:app:scaling_tiny}

\begin{table}[h]
  \centering
  \caption{\textbf{Scaling study on the QM9S test set.}
  \textbf{Top:} data scaling at a fixed 6M DiT trained for $400$k steps, sweeping the training set from $1$k to $110$k molecules.
  \textbf{Bottom:} model scaling at the fixed canonical $110$k training split; all models trained for $400$k steps.
  Within every setting we compare two settings: \emph{Struct} denoises only structure; \emph{Multi} denoises a sampled mixture of structure and spectral targets.
  Structure metrics use all-spectra conditioning. Acc@1, Comp, Atom, Bond, and Valid are percentages; both RMSD metrics are in \AA{}. Spectral cosine similarities are multiplied by $100$, while Tani\textsubscript{MG} is unscaled.
  Spec~cos columns report each spectrum generated under full cross-modal conditioning (structure $+$ the other two spectra), the strongest available regime; the strict structure-only $\rightarrow$ spectrum direction is reported in Table~\ref{tab:headline}.}
  \label{tab:scaling}
  \footnotesize
  \setlength{\tabcolsep}{3pt}
  \resizebox{\linewidth}{!}{%
  \begin{tabular}{@{}llccccccccccc@{}}
    \toprule
    & Arm
      & Acc@1{\scriptsize$\,\uparrow$}
      & RMSD\textsubscript{DS}{\scriptsize$\,\downarrow$}
      & H-RMSD{\scriptsize$\,\downarrow$}
      & Comp{\scriptsize$\,\uparrow$}
      & Atom{\scriptsize$\,\uparrow$}
      & Bond{\scriptsize$\,\uparrow$}
      & Valid{\scriptsize$\,\uparrow$}
      & Tani\textsubscript{MG}{\scriptsize$\,\uparrow$}
      & \multicolumn{3}{c}{Spec cos ($\times100$){\scriptsize$\,\uparrow$}} \\
    \cmidrule(lr){11-13}
    & & (\%) & (\AA) & (\AA) & (\%) & (\%) & (\%) & (\%) & & UV & IR & Raman \\
    \midrule
    \multicolumn{13}{@{}l}{\textit{Data scale (fixed 6M DiT)}} \\[1pt]
    1k   & Struct & $0.0$ & $1.660$ & $1.202$ & $19.7$ & $85.0$ & $90.3$ & $\mathbf{34.2}$ & $0.124$ & --- & --- & --- \\
         & Multi  & $0.0$ & \win{$1.656$} & \win{$1.164$} & \win{$24.6$} & \win{$87.5$} & \win{$91.8$} & $14.6$ & \win{$0.178$} & $90.9$ & $95.2$ & $96.8$ \\
    5k   & Struct & $0.3$ & $1.657$ & $1.074$ & $35.4$ & $89.8$ & $93.3$ & $19.1$ & $0.246$ & --- & --- & --- \\
         & Multi  & \win{$1.2$} & \win{$1.656$} & \win{$1.047$} & \win{$42.8$} & \win{$90.6$} & \win{$94.1$} & \win{$21.8$} & \win{$0.318$} & $92.0$ & $95.4$ & $96.8$ \\
    10k  & Struct & $1.6$ & $1.664$ & $0.974$ & $45.3$ & $91.0$ & $94.4$ & $25.4$ & $0.317$ & --- & --- & --- \\
         & Multi  & \win{$4.1$} & \win{$1.655$} & \win{$0.956$} & \win{$50.8$} & \win{$91.6$} & \win{$94.8$} & \win{$30.8$} & \win{$0.400$} & $92.2$ & $95.5$ & $96.8$ \\
    20k  & Struct & $9.4$ & $\mathbf{1.657}$ & $0.938$ & $\mathbf{62.6}$ & $92.2$ & $95.4$ & $39.7$ & $0.495$ & --- & --- & --- \\
         & Multi  & \win{$10.8$} & $1.662$ & \win{$0.910$} & $62.3$ & \win{$92.4$} & \win{$95.5$} & \win{$45.0$} & \win{$0.497$} & $92.4$ & $95.5$ & $96.8$ \\
    50k  & Struct & $25.4$ & $1.660$ & $0.806$ & $74.8$ & $93.6$ & $96.3$ & $59.4$ & $0.639$ & --- & --- & --- \\
         & Multi  & \win{$27.9$} & \win{$1.655$} & \win{$0.796$} & \win{$76.4$} & \win{$93.9$} & \win{$96.5$} & \win{$60.5$} & \win{$0.661$} & $92.0$ & $95.8$ & $96.7$ \\
    110k & Struct & $39.7$ & $1.662$ & $0.750$ & $\mathbf{85.3}$ & $95.1$ & $97.0$ & $66.6$ & $\mathbf{0.752}$ & --- & --- & --- \\
         & Multi  & \win{$41.5$} & \win{$1.650$} & \win{$0.745$} & $83.6$ & \win{$95.3$} & \win{$97.1$} & \win{$72.2$} & $0.738$ & $92.1$ & $95.6$ & $96.6$ \\
    \midrule
    \multicolumn{13}{@{}l}{\textit{Model scale (fixed 110k molecules)}} \\[1pt]
    S {\scriptsize(15M)}  & Struct & $51.1$ & $1.647$ & $0.608$ & $85.8$ & $95.5$ & $97.5$ & $81.5$ & $0.773$ & --- & --- & --- \\
       & Multi  & \win{$56.0$} & \win{$1.641$} & \win{$0.593$} & \win{$88.7$} & \win{$96.0$} & \win{$97.7$} & \win{$82.1$} & \win{$0.807$} & $92.5$ & $96.1$ & $97.2$ \\
    M {\scriptsize(34M)} & Struct & $55.3$ & $1.641$ & $0.551$ & $88.5$ & $95.8$ & $97.6$ & $84.1$ & $0.795$ & --- & --- & --- \\
       & Multi  & \win{$61.0$} & $1.642$ & \win{$0.522$} & \win{$91.5$} & \win{$96.3$} & \win{$97.9$} & \win{$86.5$} & \win{$0.824$} & $93.1$ & $96.5$ & $97.5$ \\
    L {\scriptsize(78M)} & Struct & $59.9$ & $1.650$ & $0.524$ & $91.2$ & $96.0$ & $97.7$ & $87.3$ & $0.813$ & --- & --- & --- \\
       & Multi  & \win{$65.4$} & \win{$1.640$} & \win{$0.486$} & \win{$93.1$} & \win{$96.5$} & \win{$98.0$} & \win{$89.9$} & \win{$0.839$} & $93.2$ & $96.6$ & $97.6$ \\
    \bottomrule
  \end{tabular}%
  }
\end{table}

With the 6M decoder, multimodal training achieves lower H-RMSD
at every training-set size from $1$k to $110$k molecules
(Table~\ref{tab:scaling}, top). The difference is
$0.038$\,\AA{} at $1$k and $0.005$\,\AA{} at $110$k.
The gains are not uniform across metrics: at $110$k, the
structure-only model achieves higher composition match and
Tani\textsubscript{MG}.

At a fixed training-set size of $110$k molecules, multimodal
training improves Acc@1 by $4.9$, $5.7$, and $5.5$ percentage
points for S, M, and L, respectively. Its H-RMSD advantage
increases from $0.015$\,\AA{} for S ($15$M) to
$0.038$\,\AA{} for L ($78$M). Spectral cosine similarity is already high with $1$k training
molecules ($95.2\%$ for IR and $96.8\%$ for Raman) and changes
little as the dataset grows. These results use structure and
the other two spectra as conditioning, a richer setting than
the structure-only conditioning evaluated in
Table~\ref{tab:headline}.

\begin{figure}[h!]
  \centering
  \includegraphics[width=0.85\linewidth]{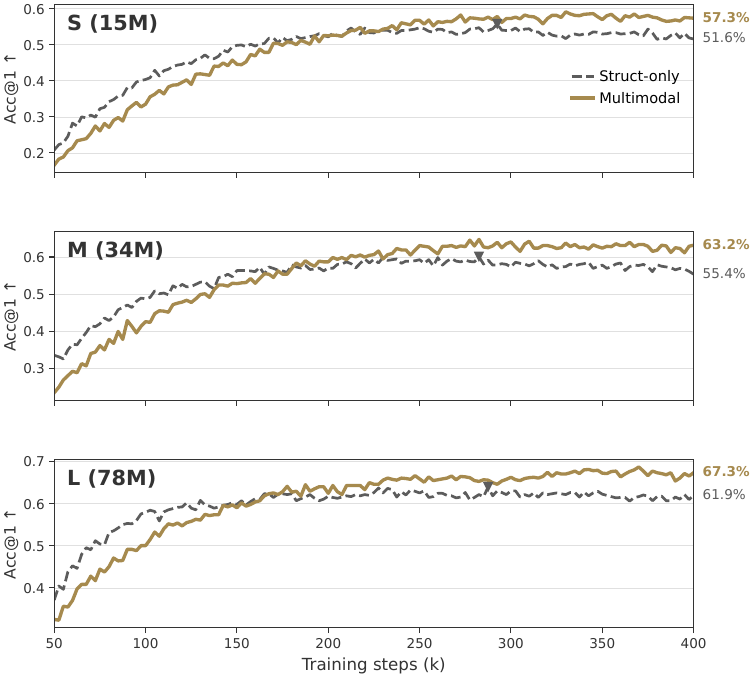}
  \caption{\textbf{Acc@1 vs.\ training step across model sizes.} Here, we investigate S, M and L model sizes. Struct-only (gray dashed) declines late in training, while multimodal (gold) remains stable or improves through $400$k steps. Triangles mark struct-only peaks.}
  \label{fig:convergence}
\end{figure}

\clearpage
Figure~\ref{fig:convergence} shows the corresponding training
curves for S, M, and L on $110$k molecules. At all three scales,
the multimodal model maintains higher Acc@1 late in training,
through $400$k training steps.

\clearpage
\section{Structure-conditioned retrieval of experimental IR spectra}
\label{sec:app:struct2ir}

We evaluate $99$ experimental gas-phase IR spectra from the EUROCHAMP
library~\citep{eurochampIR}, with one spectrum per distinct molecular
connectivity. The molecules are neutral, connected and non-radical, contain
only H, C, N, O and F, and have at most nine heavy atoms. Spectra were selected
based on molecular identity, measurement conditions, coverage and data quality,
independently of model predictions. All spectra cover $500$--$4{,}000$~cm$^{-1}$.

\paragraph{Retrieval protocol.}
Experimental spectra are interpolated onto a common wavenumber grid,
log-transformed and standardized. No correction for frequency shifts or
contamination is applied. Both models use the same RDKit-generated conformer
for each molecule. Each predicted spectrum is matched against all $99$
experimental spectra by cosine similarity. R@$k$ measures how often the
corresponding experimental spectrum appears among the top $k$ matches,
MRR is the mean reciprocal rank. All $99$ molecules are included in the results.

\paragraph{Model comparison.}
We compare MOSAIC XXL with an adapted DetaNet direct-spectrum predictor.
MOSAIC averages ten predictions per molecule in normalized spectral space,
whereas DetaNet produces one deterministic prediction. Inference settings
were fixed before evaluation at guidance scale $3$ and temperature $1$, with no
best-sample selection. 

\begin{table}[h!]
  \centering
  \caption{\textbf{Structure-conditioned retrieval on 99 experimental EUROCHAMP
  gas-phase IR spectra.} Both models query the same 99-candidate pool.
  MOSAIC averages ten predictions per molecule, DetaNet produces one
  deterministic prediction.
  R@$k$ is a percentage, MRR and cosines are dimensionless.}
  \label{tab:struct2ir}
  \small
  \setlength{\tabcolsep}{4.5pt}
  \begin{tabular}{@{}lcccccc@{}}
    \toprule
    Model & \multicolumn{2}{c}{Cosine similarity $\uparrow$}
      & R@1 $\uparrow$ & R@5 $\uparrow$ & R@10 $\uparrow$ & MRR $\uparrow$ \\
    & Median & Mean & & & & \\
    \midrule
    Random & --- & --- & $1.0$ & $5.1$ & $10.1$ & $0.052$ \\
    DetaNet & $0.130$ & $0.194$ & $5.1$ & $17.2$ & $25.3$ & $0.118$ \\
    MOSAIC & $\mathbf{0.281}$ & $\mathbf{0.308}$ & $\mathbf{15.2}$ & $\mathbf{30.3}$ & $\mathbf{39.4}$ & $\mathbf{0.242}$ \\
    \bottomrule
  \end{tabular}
\end{table}

\paragraph{Scope and limitations.}
The models are trained on simulated spectra; this experiment evaluates
retrieval of experimental spectra of molecular structures. It does not
evaluate structure generation from experimental spectra. Contaminants, mainly H$_2$O and CO, are reported for
$98$ records; spectral purity and wavenumber calibration are not established.

\paragraph{Data attribution.}
The experimental spectra are licensed under
\href{https://creativecommons.org/licenses/by/4.0/}{CC BY 4.0}, independently
of the manuscript's license. Source links, credits, provider processing and
our modifications are described in Appendix~\ref{sec:app:eurochamp_sources}.

\clearpage
\begin{figure}[h]
  \centering
  \includegraphics[width=\linewidth]{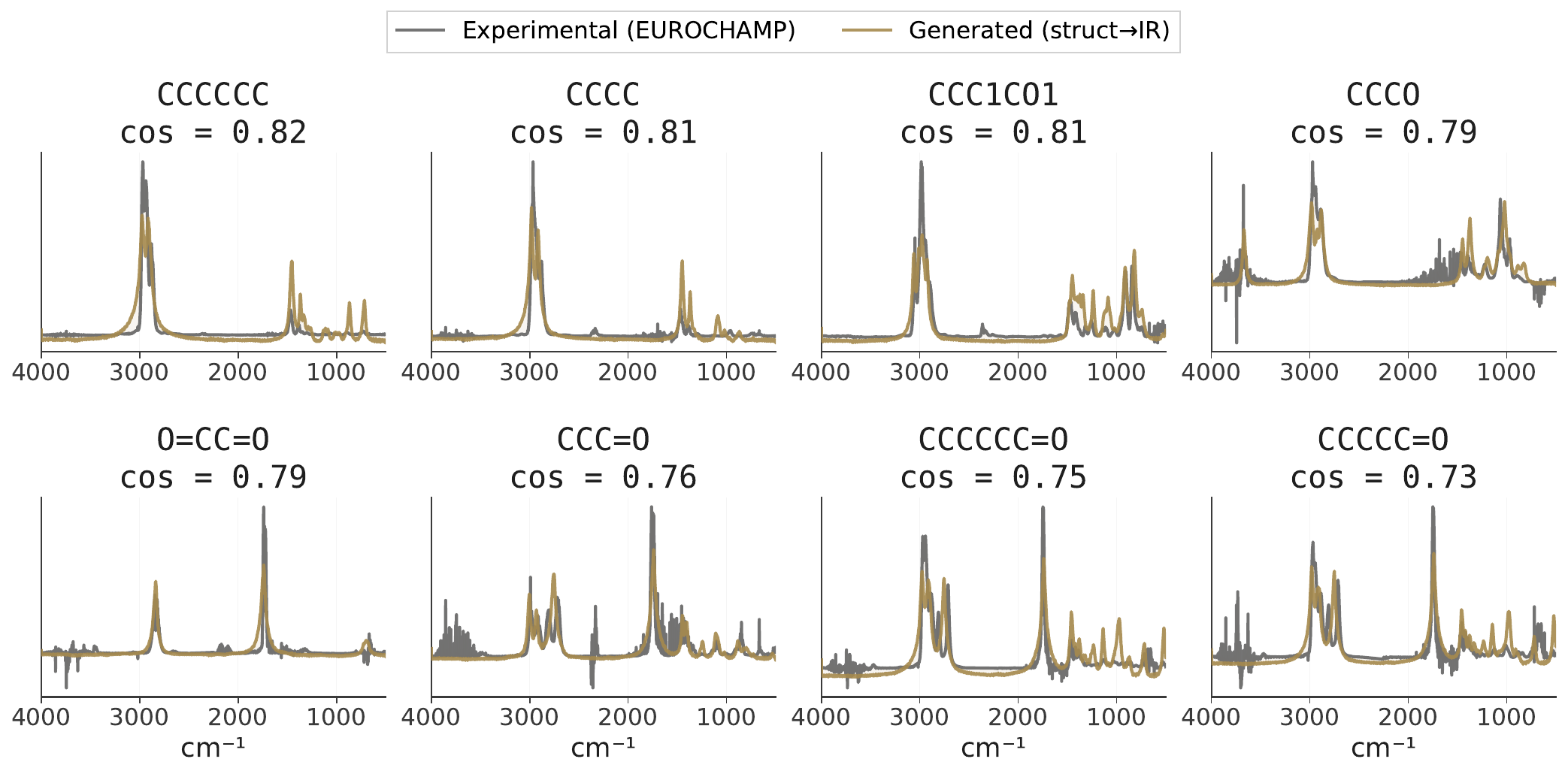}
  \caption{\textbf{Generated and experimental IR spectra.}
  MOSAIC mean predictions over ten samples (gold) and experimental spectra
  (gray) use the same model and processing as Table~\ref{tab:struct2ir}.
  Titles report cosine similarity of each plotted mean. Eight distinct molecules, each containing at most seven heavy atoms, are shown in descending cosine similarity between the mean predicted and experimental spectra. Experimental spectra are
  from EUROCHAMP~\citep{eurochampIR}, licensed CC BY 4.0. Individual sources,
  credits, provider processing and our modifications are documented in
  Appendix~\ref{sec:app:eurochamp_sources}.}
  \label{fig:struct2ir}
\end{figure}

\clearpage
\section{Spectrum-swap plots and 3D renders}
\label{sec:app:swap}

To assess whether a MOSAIC variant relies on spectral conditioning rather than a structural prior, we swap the input spectra within each of $200$ isomer pairs from the QM9S test set and regenerate molecular structures using $k\!=\!8$ samples per molecule and $50$ Euler steps.

\begin{figure}[h]
  \centering
  \includegraphics[width=\linewidth]{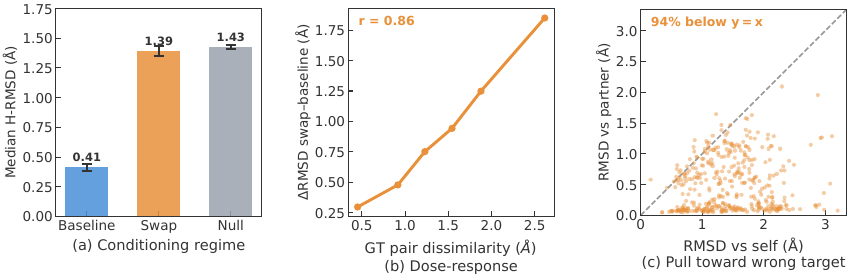}
  \caption{\textbf{Spectrum-swap intervention ($\mathbf{200}$ isomer pairs).}
    \textbf{(a)}~Median H-RMSD against the originally-paired ground truth: swap conditioning matches the no-spectrum (null) baseline, indicating the spectrum carries the conditioning signal.
    \textbf{(b)}~The swap--baseline H-RMSD gap grows linearly with the GT distance between the two pair members ($r\!=\!0.86$).
    \textbf{(c)}~For $94\%$ of the $400$ pair-sides, the median Kabsch-aligned H-RMSD over eight generated samples is lower to the partner molecule than to the originally paired ground truth.}
  \label{fig:swap}
\end{figure}

\begin{figure}[h]
  \centering
  \includegraphics[width=\linewidth]{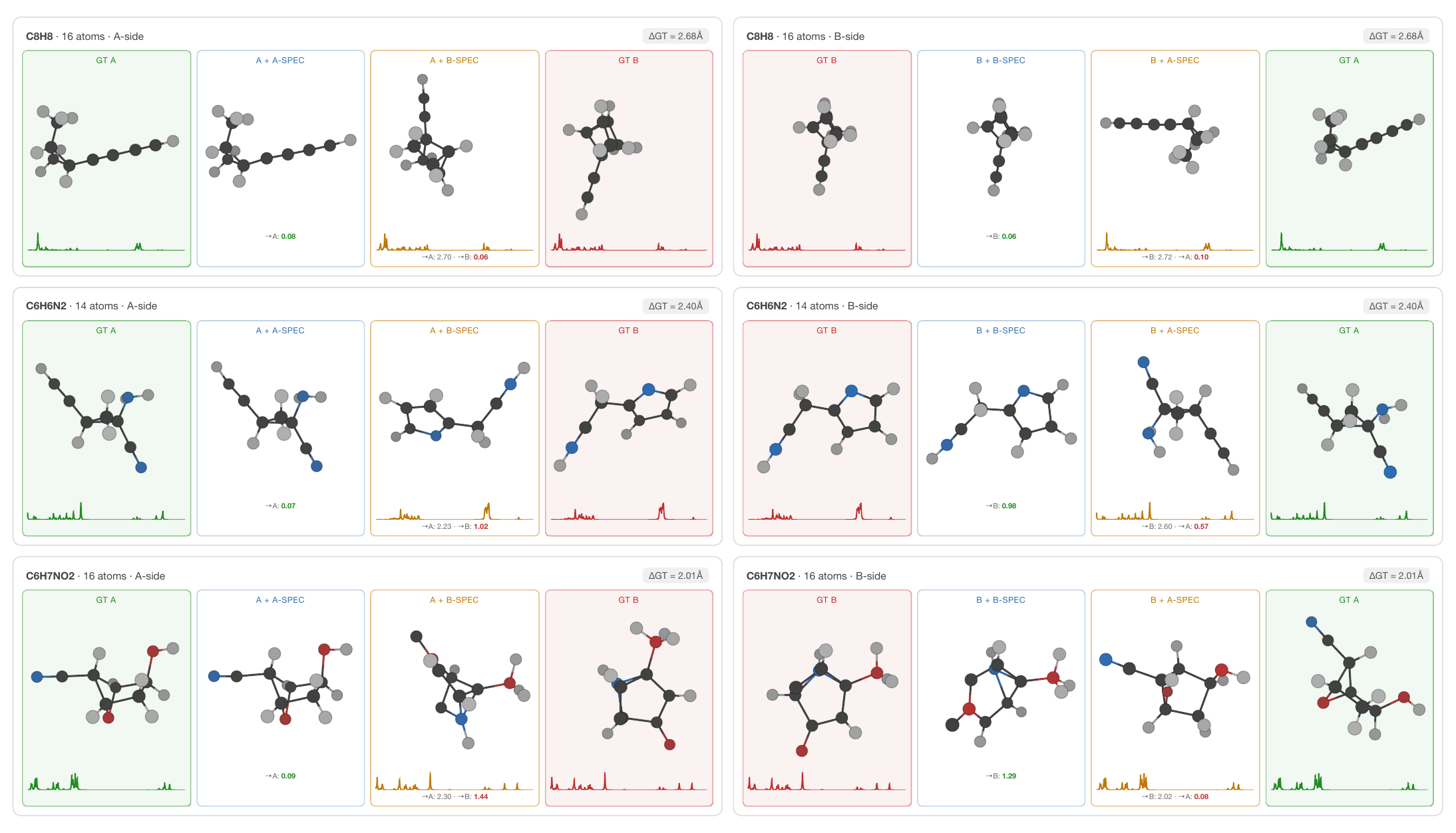}
  \caption{\textbf{Spectrum swaps for three isomer pairs.}
    Each row shows one pair, with swaps from A to B on the left and B to A on the right.
    Within each group, structures appear from left to right as the original ground truth, generation conditioned on its own spectrum, generation conditioned on the partner's spectrum, and the partner ground truth.
    Swapping the conditioning spectrum shifts the generated structure toward the partner isomer.}
  \label{fig:swap3d}
\end{figure}

\clearpage
\section{Per-modality timestep schedule}
\label{sec:app:timestep_mu}

We test whether structure and spectra benefit from different training noise distributions. Table~\ref{tab:timestep_mu} compares four schedules using a $6.2$M-parameter model trained for $100$k steps. Each run uses four transformer blocks of width $192$ and a batch size of $128$. The encoder, optimizer, and target-sampling settings are the same across runs. Only the noise distributions change.

\begin{table}[h]
  \centering
  \caption{\textbf{Effect of the training noise schedule.} $\mu$ is the mean in logit space. Larger values favor noisier inputs. Bold values mark the best result in each metric.}
  \label{tab:timestep_mu}
  \small
  \setlength{\tabcolsep}{3.5pt}
  \begin{tabular}{@{}cccccc@{}}
    \toprule
    $\mu_\textrm{struct}$
      & $\mu_\textrm{spec}$
      & H-RMSD{\scriptsize$\,\downarrow$}
      & IR cos{\scriptsize$\,\uparrow$}
      & Raman cos{\scriptsize$\,\uparrow$}
      & UV cos{\scriptsize$\,\uparrow$} \\
    \midrule
    $0$   & $0$     & \textbf{0.981}  & $0.876$          & $0.860$          & $0.794$         \\
    $+2$  & $0$     & $1.051$         & $0.878$          & $0.888$          & $0.791$         \\
    $+2$  & $-0.5$  & $1.074$         & $0.787$          & $0.739$          & $0.758$         \\
    $0$   & $+2$    & $1.011$         & \textbf{0.907}   & \textbf{0.947}   & \textbf{0.859}  \\
    \bottomrule
  \end{tabular}
\end{table}

The schedule used for MOSAIC keeps the structure mean at zero and raises the spectral mean to two. In this small-model comparison, it improves cosine similarity for all three spectra over the baseline with both means at zero, while H-RMSD increases from $0.981$ to $1.011$\,\AA{}. Raising the structure mean instead increases geometric error. Combining this change with a lower spectral mean gives the weakest results across all four metrics.

\section{Coupling timesteps across target modalities}
\label{sec:app:timestep_coupling}

The MOSAIC model in Table~\ref{tab:headline} samples training timesteps independently for each target modality. We test whether coordinating these timesteps improves generation in separate M-scale experiments.

We train three M-scale models with six transformer blocks of width $384$ for $400$k steps, using seed $42$ and global batch size $128$. All three runs use the same pretrained encoder checkpoint, data split, initialization, optimizer, and target sampler. The multimodal encoder is finetuned while UAE-3D remains frozen. For each run, we select a checkpoint on validation data and evaluate its EMA DiT weights with $50$ Euler steps and CFG $3$ on all $8{,}057$ molecules in the QM9S test set. All runs use the same inference noise with seed $0$, and joint generation uses synchronized timesteps for every target.

With \emph{independent} sampling, each target draws its own noise level using $u_m\sim\mathcal{N}(0,1)$ and $t^m=\sigma(\mu_m+u_m)$. Here $\sigma$ is the logistic sigmoid, $\mu_\mathrm{struct}=0$, and $\mu_\mathrm{spec}=2$. \emph{Shared-quantile} sampling places all simultaneous targets at the same percentile of their respective noise distributions. It draws one standard-normal $u$ per example and uses $t^m=\sigma(\mu_m+u)$, preserving each modality's noise distribution. \emph{Exact sharing} instead assigns the same timestep to every simultaneous target using $t^m=\sigma(u)$. This also removes the spectral schedule offsets, so it serves as a secondary ablation. Both coupling rules apply only to examples with at least two targets. Single-target examples retain the original sampling rule.

\begin{table}[ht]
  \centering
  \caption{\textbf{Influence of timestep coupling.} We investigate timestep coupling using M-scale experiments. Acc@1 and H-RMSD measure all-spectra$\to$structure generation.}
  \label{tab:timestep_coupling}
  \small
  \setlength{\tabcolsep}{5pt}
  \begin{tabular}{@{}lcc@{}}
    \toprule
    Timestep rule & Acc@1 (\%)$\uparrow$ & H-RMSD (\AA)$\downarrow$ \\
    \midrule
    Independent & $64.92$ & $0.579$ \\
    Shared quantile & $82.34$ & $0.359$ \\
    Exact shared & $\mathbf{82.51}$ & $\mathbf{0.355}$ \\
    \bottomrule
  \end{tabular}
\end{table}

Both coupling rules improve structure generation from spectra in these experiments. Shared quantiles also improve structure-to-spectra cosine similarity, while exact sharing reduces it. The Acc@1 gains over independent sampling are $17.41$\% with shared quantiles and $17.59$\% with exact sharing. In joint generation, shared quantiles improve mean spectral cosine by $0.063$ for UV$\to$IR+Raman+structure and $0.029$ for IR+Raman$\to$UV+structure. Shared quantiles therefore give the better balance between structure and spectral generation. In contrast, Subramanian et al. \cite{subramanian2026packflow} report that independent coordinate and lattice timesteps outperform shared timesteps in molecular crystal generation. This discrepancy could be due to a significant difference in generative tasks between the two works. In our M-scale structure–spectra experiments, coupling instead improves structure recovery, with shared-quantile sampling providing the better balance across structure and spectral metrics.

\paragraph{Gradient diagnostics.}
\label{sec:app:coupling_gradients}
Independent timesteps can mix nearly clean and heavily noised
targets within one training example. We test whether coupling
their noise levels reduces gradient variability.

We run each of the three variants for $400$k steps with one
training seed per variant. Within consecutive $50$-step windows,
we measure gradient variance before clipping, summed across all
trainable parameters (the covariance trace). We also normalize
this variance by the squared norm of the mean gradient to account
for differences in gradient magnitude.

Table~\ref{tab:coupling_gradients} summarizes $7{,}995$ paired
windows. We compute these statistics using FP64 Welford
accumulation and derive percentage changes from exponentiated
median paired log-ratios.

\begin{table}[ht]
  \centering
  \caption{\textbf{Gradient diagnostics.} Values are percentage changes relative to independent sampling. Normalized variation divides the covariance trace by the squared norm of the window-mean gradient.}
  \label{tab:coupling_gradients}
  \small
  \setlength{\tabcolsep}{5pt}
  \begin{tabular}{@{}lcc@{}}
    \toprule
    Measurement & Shared quantile & Exact shared \\
    \midrule
    Gradient variation, full run & $-5.9\%$ & $-1.3\%$ \\
    Gradient variation, first $100$k steps & $+2.8\%$ & $+4.8\%$ \\
    Gradient variation, final $100$k steps & $-14.0\%$ & $-7.6\%$ \\
    Normalized variation, full run & $+5.5\%$ & $+11.2\%$ \\
    \bottomrule
  \end{tabular}
\end{table}

Coupling does not reduce gradient variation uniformly. Both rules increase absolute variation early in training and lower it later, with a small decrease over the full run. The window-mean gradients are also smaller under coupling, so the lower absolute variation does not translate into a lower normalized measure. Non-finite gradients occur at nearly identical rates, with $170$, $169$, and $169$ events for independent, shared-quantile, and exact-shared sampling. 

\section{Training compute}
\label{sec:app:compute}

Training runs used NVIDIA H100 80\,GB GPUs. Table~\ref{tab:compute} reports approximate training costs in aggregate GPU-hours. Encoder pretraining and UAE-3D training are one-time costs reused across generative training experiments.

\begin{table}[h]
  \centering
  \caption{\textbf{Training compute.} GPU-hours are approximate. Scaling and pretraining runs used one GPU. The XXL estimate corresponds to approximately $48$ wall-clock hours on four GPUs.}
  \label{tab:compute}
  \small
  \setlength{\tabcolsep}{4pt}
  \begin{tabular}{@{}llccccc@{}}
    \toprule
    Component & Model & Hidden & Depth & Batch & Steps & GPU-h \\
    \midrule
    \multicolumn{7}{@{}l}{\textit{Pretraining}} \\[1pt]
    Encoder & large encoder & $512$ & $8$  & $128$ & $100$k & $12$ \\
    UAE-3D  & structure AE  & ---   & $6/4$ & $512$ & $430$k & $12$ \\
    \midrule
    \multicolumn{7}{@{}l}{\textit{Generative training: data scaling (6M DiT)}} \\[1pt]
    & 6M multi       & $192$ & $4$  & $128$ & $400$k & $38$ \\
    & 6M struct-only & $192$ & $4$  & $128$ & $400$k & $34$ \\
    \midrule
    \multicolumn{7}{@{}l}{\textit{Generative training: model scaling (110k data)}} \\[1pt]
    & S (15M)       & $256$ & $6$  & $128$ & $400$k & $52$ \\
    & M (34M)       & $384$ & $6$  & $128$ & $400$k & $72$ \\
    & L (78M)       & $512$ & $8$  & $128$ & $400$k & $130$ \\
    & XL (115M)     & $512$ & $12$ & $96$  & $400$k & $175$ \\
    \midrule
    \multicolumn{7}{@{}l}{\textit{Best performing model}} \\[1pt]
    & XXL (298M), unfrozen encoder & $768$ & $14$ & $256$ & $300$k & $190$ \\
    \bottomrule
  \end{tabular}
\end{table}

\section{Dataset details}
\label{sec:app:dataset}

All experiments use the QM9S dataset \citep{zou2023qm9s,molspectra2025}: QM9 molecules \citep{qm9_2014} paired with UV, IR, and Raman spectra and relaxed 3D geometries computed at the B3LYP/def-TZVP level. This is the same data source used by DiffSpectra \citep{diffspectra2025}. We use a fixed split of $110{,}000$ train / $10{,}000$ val / $8{,}057$ test molecules. Low-data experiments ($1$k--$50$k) use fixed subsets of the training set with the same val and test split. Table~\ref{tab:dataset} summarises the data representation.

\begin{table}[h]
  \centering
  \caption{\textbf{QM9S data summary.}}
  \label{tab:dataset}
  \small
  \begin{tabular}{@{}lcl@{}}
    \toprule
    Modality & Samples & Notes \\
    \midrule
    UV        & $701$     & Electronic absorption spectrum \\
    IR        & $3{,}501$ & Vibrational spectrum \\
    Raman     & $3{,}501$ & Vibrational spectrum \\
    \midrule
    Structure & \multicolumn{2}{l}{Relaxed 3D geometry, up to $29$ atoms over $\{$H, C, N, O, F$\}$.} \\
    \bottomrule
  \end{tabular}
\end{table}

\section{Encoder pretraining ablation}
\label{sec:app:phase1_ablation}

The multimodal encoder is pretrained using masked reconstruction,
cross-modal contrastive alignment, and EMA self-distillation.
We assess the contribution of the two auxiliary objectives by
removing them individually and together while retaining reconstruction.
Each configuration uses an encoder with hidden width $384$ and six
layers, trained for $50$k steps on $110$k molecules with one
training seed.

We evaluate the pretrained encoder through cross-modal reconstruction.
Four leave-one-out settings predict one modality from the other three,
while four single-modality settings predict the three missing
modalities from one observed modality. Reconstruction quality is
measured relative to the same encoder with all conditioning
modalities hidden,
$R^2_{\mathrm{PRIOR},s,m}
=1-\mathrm{MSE}_{s,m}/\mathrm{MSE}_{\mathrm{PRIOR},m}$,
where $s$ denotes the conditioning setting and $m$ the predicted
modality. Higher scores indicate greater improvement over the
baseline. This evaluation measures encoder reconstruction directly,
without training a flow matching generator.

\begin{table}[h]
  \centering
  \caption{\textbf{Effect of encoder pretraining objectives on cross-modal reconstruction.}
    The overall score averages $R^2_{\mathrm{PRIOR}}$ across all $16$ conditioning--target pairs. LOO and ONE report averages over the four leave-one-out and twelve single-modality pairs, respectively. Per-target columns average the four pairs predicting each modality. Higher scores are better.}
  \label{tab:phase1_knockout}
  \small
  \setlength{\tabcolsep}{4pt}
  \begin{tabular}{@{}lccc ccc cccc@{}}
    \toprule
    & \multicolumn{3}{c}{Loss weights}
      & \multicolumn{3}{c}{Aggregate $R^2_\mathrm{PRIOR}$}
      & \multicolumn{4}{c}{Per-target $R^2_\mathrm{PRIOR}\!\uparrow$} \\
    \cmidrule(lr){2-4} \cmidrule(lr){5-7} \cmidrule(l){8-11}
    Config
      & $\lambda_\mathrm{recon}$
      & $\lambda_\mathrm{contr}$
      & $\lambda_\mathrm{distill}$
      & Score $\!\uparrow$
      & LOO $\!\uparrow$
      & ONE $\!\uparrow$
      & struct
      & UV
      & IR
      & Raman \\
    \midrule
    Full            & $1.0$ & $0.5$ & $0.5$ & $.551$ & $.751$ & $.484$ & $.344$ & $.620$ & $.617$ & $.621$ \\
    No contrastive  & $1.0$ & $0.0$ & $0.5$ & $.545$ & $.749$ & $.476$ & $.343$ & $.612$ & $.609$ & $.614$ \\
    No distillation & $1.0$ & $0.5$ & $0.0$ & $.551$ & $.752$ & $.484$ & $.344$ & $.619$ & $.618$ & $.623$ \\
    Recon only      & $1.0$ & $0.0$ & $0.0$ & $.544$ & $.747$ & $.477$ & $.342$ & $.609$ & $.610$ & $.616$ \\
    \bottomrule
  \end{tabular}
\end{table}

Reconstruction alone achieves an aggregate score of $0.544$,
compared with $0.551$ for the full objective.
Adding contrastive alignment improves the score by
$0.006$--$0.007$, with gains across all four target modalities
both with and without distillation.
Self-distillation has little effect on the aggregate reconstruction
score and produces mixed changes across modalities.
At this model scale and training budget, the single-seed results
show modest gains in prior-relative reconstruction scores from
contrastive alignment, with little additional benefit from
self-distillation. Its contribution to downstream structure
generation is examined next.

\paragraph{Distillation targets.}
The reconstruction scores above do not directly measure whether distillation helps the encoder support structure generation.
We therefore compare three distillation targets through downstream generation, alongside a model trained without distillation.
Full-sequence mean pooling averages all student and teacher states, including CLS tokens, padded atom slots, and dropped-modality positions.
CLS-only pooling averages the four modality CLS states.
Tokenwise distillation matches corresponding student and teacher states, excludes padded atoms and dropped modalities, and weights each observed modality equally.
All three use cosine distance between projected representations, with reconstruction and contrastive alignment unchanged.

For this study, we pretrain multimodal encoders with eight blocks of width $512$ for $200$k steps and select checkpoints by validation reconstruction loss.
Each encoder is then frozen while a matched M-scale generative model is trained for $400$k steps.
We repeat the comparison with three training seeds and evaluate spectra-to-structure generation on the full $8{,}057$-molecule QM9S test split.

\begin{table}[h]
  \centering
  \caption{\textbf{Influence of distillation targets.} Evaluated through structure generation with frozen encoders. Values are mean $\pm$ standard deviation across three training seeds.
    Bold values mark the best mean in each metric.}
  \label{tab:distillation_targets}
  \small
  \setlength{\tabcolsep}{8pt}
  \begin{tabular}{@{}lccc@{}}
    \toprule
    Distillation target & Acc@1 ($\%$) $\uparrow$ & H-RMSD (\AA) $\downarrow$ & Composition ($\%$) $\uparrow$ \\
    \midrule
    Full-sequence mean & $58.87 \pm 3.95$ & $0.548 \pm 0.075$ & $90.19 \pm 0.72$ \\
    CLS-only           & $63.00 \pm 0.14$ & $0.496 \pm 0.008$ & $\mathbf{91.28} \pm 0.69$ \\
    Tokenwise          & $\mathbf{63.04} \pm 2.10$ & $\mathbf{0.484} \pm 0.013$ & $90.47 \pm 1.05$ \\
    No distillation    & $61.59 \pm 3.16$ & $0.504 \pm 0.023$ & $90.03 \pm 1.32$ \\
    \bottomrule
  \end{tabular}
\end{table}

\begin{samepage}
Tokenwise distillation gives the lowest mean H-RMSD in Table~\ref{tab:distillation_targets}, while CLS-only achieves similar mean graph accuracy and the highest composition match.
Full-sequence mean pooling gives the weakest average graph accuracy and geometry.
Neither alternative consistently improves on no distillation across seeds.

\end{samepage}

\section{Encoder training and multimodal generation}
\label{sec:app:attribution}

We compare three encoder training strategies: a frozen pretrained
encoder, a pretrained encoder fine-tuned jointly with the generator,
and a randomly initialized encoder trained jointly with the generator.
For each strategy, we compare multimodal and structure-only training
using an M-scale generator with hidden width $384$, six layers,
and a training budget of $400$k steps. These results compare
complete training procedures and do not isolate the effect of encoder
fine-tuning.

\begin{table}[h]
  \centering
  \caption{\textbf{Encoder training and multimodal generation.}
Spectra-to-structure results for three encoder training strategies
under multimodal and structure-only objectives. $\Delta$Acc@1
reports the multimodal gain within each strategy, in percentage points.}
  \label{tab:attribution}
  \small
  \setlength{\tabcolsep}{4pt}
  \begin{tabular}{@{}llccccc@{}}
    \toprule
    Encoder & Targets & Acc@1{\scriptsize$\,\uparrow$} & H-RMSD{\scriptsize$\,\downarrow$} & Comp{\scriptsize$\,\uparrow$} & Tani\textsubscript{MG}{\scriptsize$\,\uparrow$} & $\Delta$Acc@1 \\
    \midrule
    Frozen pretrained         & Multi       & $60.2$ & $0.538$ & $90.9$ & $0.820$ &  $+6.8$\\
    Frozen pretrained         & Struct-only & $53.4$ & $0.553$ & $86.8$ & $0.790$ & \\
    \addlinespace
    Random-init               & Multi       & $72.5$ & $0.449$ & $94.8$ & $0.883$ & $+5.3$ \\
    Random-init               & Struct-only & $67.2$ & $0.468$ & $92.5$ & $0.863$ &  \\
        \addlinespace
    Pretrained, fine-tuned    & Multi       & $\mathbf{85.4}$\textsuperscript & $0.291$\textsuperscript & $\mathbf{97.4}$\textsuperscript & $\mathbf{0.944}$\textsuperscript & $+1.8$\\
    Pretrained, fine-tuned    & Struct-only & $83.6$ & $\mathbf{0.280}$ & $96.3$ & $0.937$ &  \\
    \bottomrule
  \end{tabular}

\end{table}

The pretrained, fine-tuned models achieve the highest Acc@1
under both target regimes in Table~\ref{tab:attribution}.
Their scores exceed those of the frozen encoders by
$25.2$--$30.2$\% and those of the randomly
initialized encoders by $12.9$--$16.4$\%.

Within each encoder strategy, multimodal training achieves
higher Acc@1 than structure-only training, with gains of
$6.8$, $5.3$, and $1.8$\% for frozen pretrained,
randomly initialized, and pretrained fine-tuned encoders,
respectively. The geometry results are mixed: multimodal training
reduces H-RMSD in the frozen and randomly initialized settings,
but has slightly higher H-RMSD in the pretrained fine-tuned
setting ($0.291$ versus $0.280$\,\AA{}).

\begin{figure}[h]
  \centering
  \includegraphics[width=0.6\linewidth]{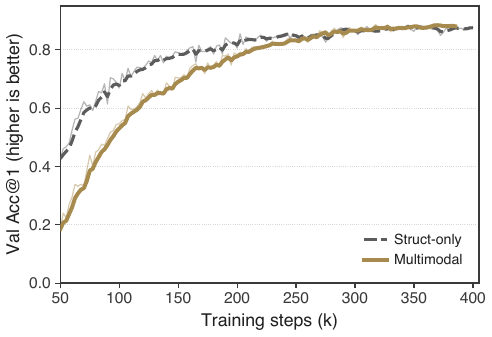}
  \caption{\textbf{Validation accuracy during joint encoder and generator training.}
Acc@1 for the pretrained fine-tuned M-scale models under
multimodal and structure-only objectives.}
  \label{fig:attribution_convergence}
\end{figure}

Figure~\ref{fig:attribution_convergence} shows validation Acc@1
for the pretrained finetuned models. The multimodal model
initially trails the structure-only model and reaches similar
accuracy later in training.

\section{EUROCHAMP data attribution}
\label{sec:app:eurochamp_sources}
The experimental spectra in Figure~\ref{fig:struct2ir} are adapted from the
EUROCHAMP collection published by AERIS under
CC BY 4.0.
We credit Mila R\'odenas and, for 1,2-epoxybutane, Iulia Patroescu-Klotz and
Ian Barnes; the hexanal measurements were made by H.~Plagens at EUPHORE.
The individual sources, in figure reading order, are \href{https://doi.org/10.25326/ffnn-m774}{hexane},
\href{https://doi.org/10.25326/w7qj-zb16}{butane}, \href{https://doi.org/10.25326/f9pp-1270}{1,2-epoxybutane},
\href{https://doi.org/10.25326/g7vq-n828}{1-propanol}, \href{https://doi.org/10.25326/ecnh-5582}{biformile},
\href{https://doi.org/10.25326/emqn-sm52}{propanal}, \href{https://doi.org/10.25326/3kma-bh67}{hexanal}, and
\href{https://doi.org/10.25326/az6q-cn27}{pentanal}. The source spectra had undergone background referencing, conversion to
absorbance, truncation and scaling, with additional arithmetic and
straight-line processing documented in the linked datasets.
We resample and normalize these spectra and overlay model predictions as
described in Appendix~\ref{sec:app:struct2ir}. The reproduced spectral material
remains under CC BY 4.0, independently of the paper's license.

\end{document}